\PassOptionsToPackage{colorlinks=true,linkcolor=blue,citecolor=blue,urlcolor=blue}{hyperref}

\documentclass[sigconf]{acmart}

\usepackage{graphicx}
\usepackage{amsmath}
\usepackage{makecell}
\usepackage{booktabs}
\usepackage{algorithm}
\usepackage{algorithmic}
\usepackage{url}
\usepackage{multirow}
\usepackage{graphicx}
\usepackage{epstopdf}
\usepackage{booktabs}
\usepackage{amsthm}
\newtheorem{remark}{Remark}
\newtheorem{lemma}{Lemma}

\newtheorem{theorem}{Theorem}

\usepackage{enumitem}
\usepackage{array}
\usepackage{subfigure}
\usepackage{color}
\usepackage{wasysym}
\usepackage{xcolor}  
\usepackage{tikz}  
\usepackage{marvosym}

\newcommand{\mrone}[1]{\textcolor{black}{#1}}
\newcommand{\rone}[1]{\textcolor{black}{#1}}
\newcommand{\rtwo}[1]{\textcolor{black}{#1}}
\newcommand{\rthree}[1]{\textcolor{black}{#1}}

\definecolor{darkgreen}{RGB}{37, 193, 123}
\definecolor{whitepink}{RGB}{237,0,140}

\AtBeginDocument{%
  
    }

\setcopyright{acmlicensed}
\copyrightyear{2026}
\acmYear{2026}
\acmDOI{XXXXXXX.XXXXXXX}
\acmConference{2026 ACM SigMOD/PODS International Conference on Management of Data}{May 31-June 5, 2026,
  }{Bangalore, India}
\acmISBN{978-1-4503-XXXX-X/2018/06}

\acmSubmissionID{205}

\setcopyright{acmlicensed}
\copyrightyear{2026}
\acmYear{2026}
\acmDOI{XXXXXXX.XXXXXXX}

\begin{document}

\title{Categorical Data Clustering via Value Order Estimated \\Distance Metric Learning}

\author{Yiqun Zhang}
\affiliation{%
  \institution{School of Computer Science and Technology, Guangdong University of Technology, Guangzhou, China}
  \city{}
  \country{}
}
\authornote{Co-first author}
\email{yqzhang@gdut.edu.cn}

\author{Mingjie Zhao}
\affiliation{%
  \institution{Department of Computer Science, \\ Hong Kong Baptist University,}
  \city{Hong Kong SAR}
  \country{China}
}
\authornotemark[1] 
\email{mjzhao@life.hkbu.edu.hk}

\author{Hong Jia}
\authornote{Corresponding author}
\affiliation{%
  \institution{Guangdong Provincial Key Laboratory of Intelligent Information Processing, College of Electronics and Information Engineering, Shenzhen University, Shenzhen, China}
  \city{}
  \country{}
}
\email{hongjia1102@szu.edu.cn}

\author{Mengke Li}
\affiliation{%
  \institution{School of Computer Science and Software Engineering, Shenzhen University, Shenzhen, China}
  \city{}
  \country{}
}
\email{mengkeli@szu.edu.cn}

\author{Yang Lu}
\affiliation{%
  \institution{School of Informatics, \\ Xiamen University,}
  \city{Xiamen}
  \country{China}
}
\email{luyang@xmu.edu.cn}

\author{Yiu-ming Cheung}
\authornotemark[2]
\affiliation{%
  \institution{Department of Computer Science, \\ Hong Kong Baptist University,}
  \city{Hong Kong SAR}
  \country{China}
}
\email{ymc@comp.hkbu.edu.hk}

\begin{abstract}
Clustering is a popular machine learning technique for data mining that can process and analyze datasets to automatically reveal sample distribution patterns. Since the ubiquitous categorical data naturally lack a well-defined metric space such as the Euclidean distance space of numerical data, the distribution of categorical data is usually under-represented, and thus valuable information can be easily twisted in clustering.
This paper, therefore, introduces a novel order distance metric learning approach to intuitively represent categorical attribute values by learning their optimal order relationship and quantifying their distance in a line similar to that of the numerical attributes. Since subjectively created qualitative categorical values involve ambiguity and fuzziness, the order distance metric is learned in the context of clustering. Accordingly, a new joint learning paradigm is developed to alternatively perform clustering and order distance metric learning with low time complexity and a guarantee of convergence. 
\rthree{Due to the clustering-friendly order learning mechanism and the homogeneous ordinal nature of the order distance and Euclidean distance, the proposed method achieves superior clustering accuracy on categorical and mixed datasets. More importantly, the learned order distance metric greatly reduces the difficulty of understanding and managing the non-intuitive categorical data.}
Experiments with ablation studies, significance tests, case studies, etc., have validated the efficacy of the proposed method. The source code is available at \href{https://github.com/DAJ0612/OCL_Source_Code}{{https://github.com/DAJ0612/OCL\_Source\_Code}}.

\end{abstract}



\maketitle
\section{Introduction}

Categorical attributes with qualitative values (e.g., $\{$lawyer, doctor, computer scientist$\}$ of an attribute ``occupation'') are common in cluster analysis of real datasets. As the categorical values are cognitive conceptual entities that cannot directly attend arithmetic operations~\textcolor{blue}{\cite{intro1, CDS_VLDBJ}}, most existing clustering techniques proposed for numerical data are not directly applicable to the qualitative-valued categorical data~\textcolor{blue}{\cite{intro4,intro3}}. Although some conventional encoding strategies, such as one-hot encoding, are feasible to convert categorical data into numerical data, information loss is often inevitable. Therefore, the development of categorical data clustering techniques, including clustering algorithms, distance measures, and distance metric learning approaches, continues to attract attention in data management and machine learning fields~\textcolor{blue}{\cite{intro10, sccd, Geccb}}.
Categorical data clustering approaches can be roughly categorized into: 1) $k$-modes and its variants~\textcolor{blue}{\cite{CDCDR,mwkm,woc}} that partition data samples into clusters based on certain distance measures, and 2) cluster quality assessment-based algorithms~\textcolor{blue}{\cite{ICDCSCai,sbcc,ecc}} that adjust the sample-cluster affiliations based on certain cluster quality measures. Their performance is commonly dominated by the adopted metric space. Thus, the focus of current research has shifted from the design of clustering algorithms to proposing reasonable measures that suit categorical data clustering~\textcolor{blue}{\cite{intro7}}.

\begin{figure}[!t]	
\centerline{\includegraphics[width=3.3in]{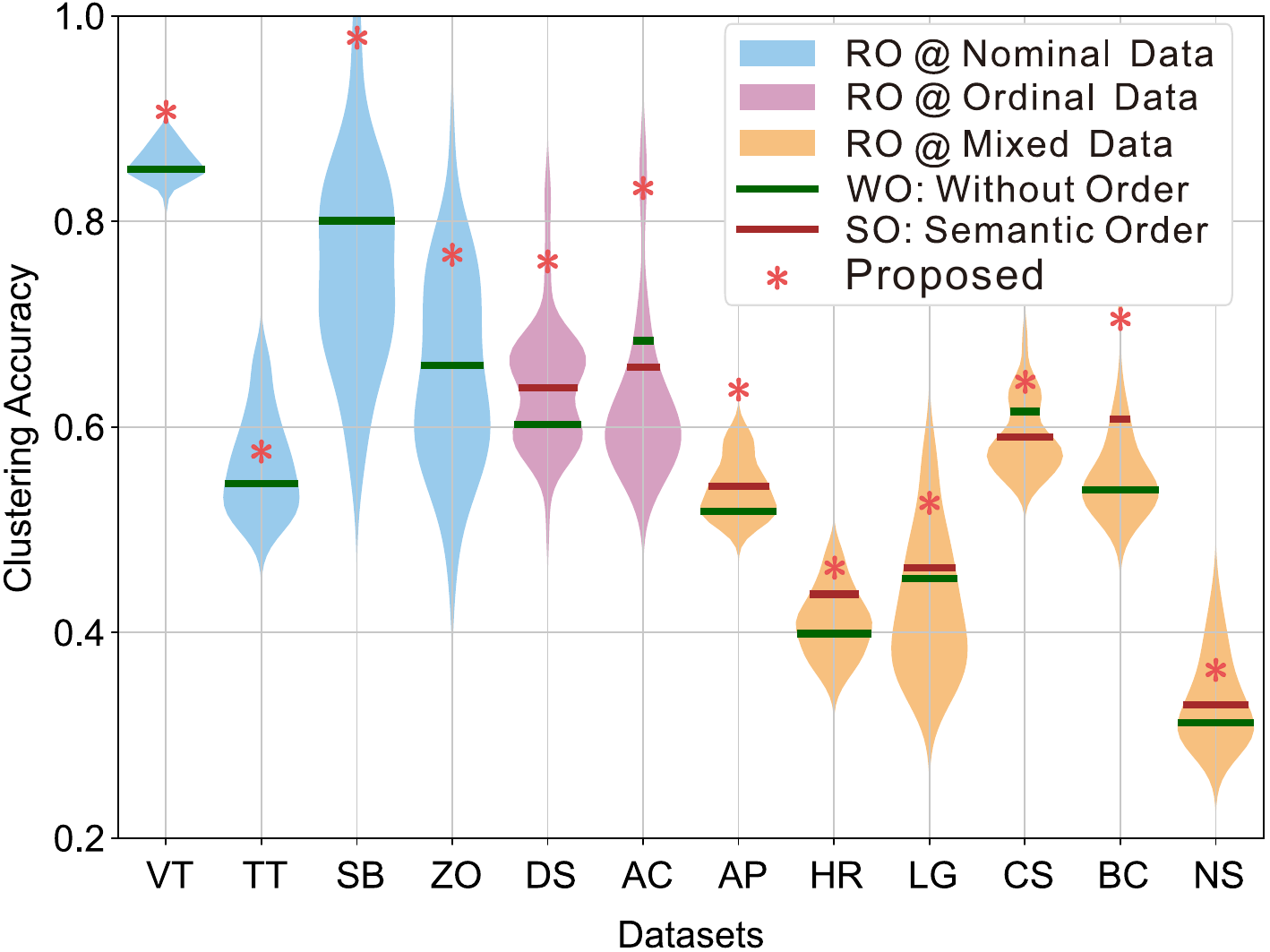}}
\caption{\rtwo{Clustering accuracy of our method (indicated by red stars)} and the variants of $k$-modes~\textcolor{blue}{\cite{kmd}}: 1) WO: Without considering semantic Orders indicated by green bars, 2) SO: with Semantic Orders indicated by dark red bars, and 3) RO: with Random Orders generated 1000 times to characterize the clustering accuracy distribution on datasets comprising nominal attributes (blue region), ordinal attributes (purple region), and both (i.e., mixed data indicated by brown region). 
}
\label{fig:pre_ex}	
\end{figure}

The Hamming distance and one-hot encoding are common measure and encoding strategy, respectively, and they both simply represent the distance between two categorical values based on whether the values are identical or not, which is far from informative. Later, distance measures have been proposed to finely indicate the differences at the value level by considering the intra-attribute value probabilities~\textcolor{blue}{\cite{emd3}}, the inter-attribute dependence~\textcolor{blue}{\cite{adm,cbdm_journal,abdm}}, or both~\textcolor{blue}{\cite{jdm,udm}}. Advanced encoding strategies that encode distances between data samples~\textcolor{blue}{\cite{sbc,GFBD}}, and couplings of attribute values~\textcolor{blue}{\cite{cde_conf,cde,QGRL}} have also been proposed. Since they are clustering task independent,
learning-based approaches have been further developed to optimize the representation and clustering jointly to circumvent suboptimal solutions. Typical learning-based works include the distance learning-based methods~\textcolor{blue}{\cite{oc,woc,ADC}} that learn the dissimilarity relationship of categorical values, and representation learning-based methods~\textcolor{blue}{\cite{COForest,HARR,untie}} that learn to project categorical attribute values into a higher-dimensional Euclidean distance space.

\rtwo{\rone{Current distance or representation learning approaches typically involve dimensional expansion of categorical attributes, which impedes efficient data management~\rone{\cite{SIGMOD1,MICCF_TKDD24}}}. Furthermore, when processing mixed datasets comprising both categorical and numerical attributes, these approaches inadvertently amplify the weight of categorical attributes. This introduces heterogeneity and thus compromises the reliability of cluster analysis results. To address this, some methods project categorical attribute values into one-dimensional Euclidean distance metric space based on their semantic order~\textcolor{blue}{\cite{iNDIA24pr,HDC}} (e.g., an attribute ``decision'' with values $\{$strong\_accept, accept, \dots, strong\_reject$\}$), aiming to align them with numerical attributes. However, this strategy is inapplicable to attributes lacking semantic order. Even when a semantic order exists, it is often subjectively assigned by the data collector. Consequently, these methods demonstrate significant clustering performance improvements only on datasets where the semantic order is highly relevant to the clustering task. This motivates a more general \textit{hypothesis} regarding the concept of ``order'': \textit{The implicit optimal order of categorical attribute values (even in the absence of semantic order) can substantially enhance clustering performance.}
}

To validate the hypothesis, a demonstration experiment (illustrated in Figure~\textcolor{blue}{\ref{fig:pre_ex}}) is conducted on twelve datasets, with statistical characteristics detailed in Table~\textcolor{blue}{\ref{tb:detaile_infor}} within Section~\textcolor{blue}{\ref{sct:exp_Set}}. The standard $k$-modes~\textcolor{blue}{\cite{kmd}} clustering algorithm is employed adopting the following three distance metrics: 1) the original Hamming distance, i.e., treat categorical attributes Without Orders (WO); 2) Semantic Order (SO) distance, i.e., treat ordinal attribute values like $\{$accept, neutral, reject$\}$ with normalized order distances $dist(\text{accept},\text{neutral})=0.5$, $dist(\text{neutral},$ $\text{reject})=0.5$, $dist(\text{accept},$ $\text{reject})=1$; and 3) Random Order (RO) distance, i.e., distances derived from randomly assigned orders to the values of each attribute, followed by normalization. Note that SO is inapplicable to the left four datasets, which consist solely of nominal attributes. Due to the random initialization of cluster modes, $k$-modes+WO and $k$-modes+SO are each executed 100 times, with average performance reported. $k$-modes+RO is implemented 1000 times to adequately characterize the distribution of clustering accuracy. Results indicate that although SO is utilized on the right 8 datasets, its performance (red bars) does not consistently surpass that of WO (green bars). This occurs because the semantic order, being subjectively defined by data collectors, may not align with the clustering task. Crucially, the 1000 trials of RO adequately traverse the state space of implicit value orders of the attributes. Thus, the hypothesis is strongly validated by the observation that the upper performance limits of RO are markedly higher than the performance of SO and WO. Therefore, the above evidence suggests that identifying the implicit optimal orders represents a critical factor for achieving accurate categorical data clustering.

This paper, therefore, proposes a new order learning-based clustering paradigm that dynamically estimates orders based on the current data partition to optimize the data partition. To make full use of the learned orders, a distance learning mechanism is also introduced to dynamically update the order distance measure for the sample-cluster affiliation. Since the order learning and the clustering task are connected to form an integrated optimization problem, the sub-optimal clustering solutions can be considerably circumvented. It turns out that the learned orders can significantly improve clustering accuracy and can also intuitively help understand the distribution and clustering effect of categorical data. The convergence of the proposed optimization algorithm and the metric properties of the order distance are rigorously guaranteed. Extensive experiments, including comparative studies, ablation studies, significance tests, case studies, and intuitive visualization, demonstrate the superiority of the proposed paradigm.
The main contributions of this work can be summarized into the following three points:
\begin{itemize}
    \item A new insight that ``obtaining the optimal orders is the decisive factor for accurately clustering categorical data'' is proposed. Accordingly, a new clustering paradigm with order distance learning is presented.
    \item The proposed iterative learning mechanism more thoroughly optimizes the order, distance, and clusters by considering their connections. As a result, it features adaptability and is more robust to various clustering scenarios.
    \item Both the learning process and learned orders are highly interpretable, which helps understand complex categorical data. Moreover, the algorithm is efficient, strictly convergent, and can be easily adapted for mixed data clustering.
\end{itemize}

\section{Related Work}

This section overviews the related works in: 1) \textbf{data encoding}, 2) \textbf{representation learning}, and 3) \textbf{consideration of value order}, from the perspective of categorical data clustering.

The \textbf{data encoding} techniques convert categorical data into numerical values based on certain strategies and then cluster numerical encoding. As the conventional one-hot encoding and the Hamming distance cannot distinguish different degrees of dissimilarity~\textcolor{blue}{\cite{intro6,intro10}} of data samples, information loss is usually unavoidable. \rone{Some statistic-based distance measures~\textcolor{blue}{\cite{oc,lsm,CDS_VLDBJ}} consider the intra-attribute occurrence frequency of values to better distinguish the data samples.} Some metrics~\textcolor{blue}{\cite{adm,cbdm_journal,abdm}} further define distance considering the relevance between attributes. Later, the measures~\textcolor{blue}{\cite{jdm,cms}} and encoding strategies~\textcolor{blue}{\cite{cde_conf,cde,sbc}} that more comprehensively consider both the intra- and inter-attribute couplings reflected by the data statistics have also been presented. 

Recently, more advanced \textbf{representation learning} methods have been proposed to learn categorical data encoding, facilitating better data representation in data analysis tasks. The work proposed in~\textcolor{blue}{\cite{untie}} projects dataset through different kernels for more informative coupling representation, and then learns the kernel parameters to obtain task-oriented representation. In contrast, \textcolor{blue}{\cite{ICDCSCai}} reveals the complex nested multi-granular cluster effect to obtain multi-granular distribution encoding representation of categorical data. \rone{Some approaches~\textcolor{blue}{\cite{CaFE_TKDD25,SIGMOD2,mix2vec}} have also been designed to simultaneously encode numerical and categorical attributes. However, their performance is highly sensitive to hyper-parameters, making optimal selection labor-intensive across diverse real-world applications.} Although a parameter-free approach~\textcolor{blue}{\cite{woc}} has been proposed to learn sample-cluster distance based on cluster statistics, it does not form a valid distance metric for categorical data.

Some new advances proposed with the \textbf{consideration of value order} further distinguish categorical attributes into nominal and ordinal attributes in clustering tasks. The distance metrics~\textcolor{blue}{\cite{ ebdmjournalLLL, udm}} define the distance between attribute values based on the data statistical information with considering the semantic order. Later, a representation learning approach was introduced by~\textcolor{blue}{\cite{dlc}}, which treats the possible values of an ordinal attribute as linearly arranged concepts, and learns the distances between adjacent concepts. Later, the works~\textcolor{blue}{\cite{ADC,HDC}} further unify the distances of nominal and ordinal attributes, and make them learnable with clustering. The most recent work~\textcolor{blue}{\cite{het2hom,HARR}} attempts to convert the nominal attributes into ordinal ones through geometric projection, and then learn the latent distance spaces during clustering. The above recent advances achieve considerable clustering performance improvements and empirically prove the importance of value order in categorical data clustering. However, they all assume the availability of the semantic order of attribute values.
Moreover, they directly adopt this order for clustering, and overlook that the semantic order may not be suitable under different clustering scenarios as shown in Figure~\textcolor{blue}{\ref{fig:pre_ex}}.

\section{Proposed Method}
\label{sct:PM}
This section first formulates the problem, then proposes the order distance definition, order learning mechanism, and the joint learning of the order-guided distances and clusters with analysis.

\subsection{Problem Formulation}

\rthree{The problem of clustering data with categorical attributes is formulated as follows. Given a dataset $X=\{\mathbf{x}_1,\mathbf{x}_2,...,\mathbf{x}_n\}$ with $n$ samples (e.g., a collection of $n$ clients of a bank). Each sample $\mathbf{x}_i$ can be denoted as an $s$-dimensional row vector $\mathbf{x}_i=[x_{i,1},x_{i,2},..,x_{i,s}]^\top$ represented by $s$ attributes (e.g., occupation, priority, income, etc.) including $s^{\{c\}}$ categorical and $s^{\{u\}}$ numerical attributes with $s^{\{c\}}+s^{\{u\}}=s$. For simplicity and without loss of generality, assume that all categorical and numerical attributes come from the corresponding attribute sets $A^{\{c\}}$ and $A^{\{u\}}$, respectively. Each attribute can be denoted as a column vector $\mathbf{a}_r=[x_{1,r},x_{2,r},...,x_{n,r}]$ composed of the description of all $n$ samples, all taking values from a corresponding possible value set $V_r=\{v_{r,1},v_{r,2},...,v_{r,l_r}\}$.}

\rthree{For a categorical attribute $\mathbf{a}_r\in A^{\{c\}}$, the possible value set is qualitative, e.g., $\{\text{lawyer}, \text{doctor}, ..., \text{scientist}\}$ of an ``occupation'' attribute, while for a numerical attribute $\mathbf{a}_r\in A^{\{u\}}$, the value set is quantitative, e.g., $\{\$4228, \$11900, ..., \$2880\}$ of an ``income'' attribute. We use $l_r$ to indicate the total number of possible values of an attribute, and for a typical categorical attribute, $l_r$ is usually finite and satisfies $l_r<n$. Note that the subscript sequential numbers of possible values only distinguish them, and do not indicate their order. We do not distinguish between nominal and ordinal attributes under categorical attributes in the whole Section~\textcolor{blue}{\ref{sct:PM}}, because their order will be learned uniformly by the proposed method.}

\rthree{Like numerical attributes with inherent value order, there is also an optimal order indicating an appropriate underlying distance metric of each categorical attribute $\mathbf{a}_r$. The difference is that there exists ambiguity in the relationship among categorical possible values, so their potential optimal order should be learned in a specific context, i.e., clustering here.} We denote the optimal order of $\mathbf{a}_r$ as $O^*_r=\{o^*(v_{r,1}),o^*(v_{r,2}),...,o^*(v_{r,l_r})\}$ where the superscript ``$*$'' indicates the optimum and the value of $o^*(v_{r,g})$ is an integer reflecting the unique ranking of $v_{r,g}$ among all its $l_r$ possible values. \rthree{For a numerical $\mathbf{a}_r\in A^{\{u\}}$, since the attribute values themselves already precisely reflect their orders in Euclidean distance space, we have $o(v_{r,g})=v_{r,g}$ with $g=\{1,2,...,l_r\}$, and the Euclidean distance $\text{dist}(v_{r,g},v_{r,h})=|o(v_{r,g})-o(v_{r,h})|=|v_{r,g}-v_{r,h}|$ can be viewed as reflected by the order of values.} For a categorical $\mathbf{a}_r\in A^{\{c\}}$, by still taking the ``occupation'' attribute as an example, if its three possible values $v_{r,1}=\text{lawyer}$, $v_{r,2}=\text{doctor}$, and $v_{r,3}=\text{scientist}$, ranking second, third, and first, respectively, then the corresponding order is $O_r=\{o(\text{lawyer}),o(\text{doctor}),o(\text{scientist})\}=\{2,3,1\}$, which reflects value relationship satisfying: $\text{dist}(v_{r,g},v_{r,h})\propto |o(v_{r,g})-o(v_{r,h})|$.

\begin{table}[!t]
\caption{Frequently used symbols. Lowercase, uppercase, bold lowercase, and bold uppercase are uniformly used to indicate value, set, vector, and matrix, respectively.}
\label{tb:symbol}
\centering
\resizebox{1\columnwidth}{!}{
\begin{tabular}{c|l}
\toprule
Symbol & Explanation  \\
\midrule
$X$, $A^{\{c\}}$, and $A^{\{u\}}$& Dataset, categorical attribute set, and numerical attribute set \\	
$O$ and $C$& order set corresponding to categorical attributes and cluster set \\	
$\mathbf{x}_i$ and $x_{i,r}$& $i$-th data object and $r$-th value of $\mathbf{x}_i$  \\	
$\mathbf{a}_r$, $V_r$, and $v_{r,g}$& $r$-th attribute, its value set, and $g$-th value of $V_r$\\	
$l_r$& Total number of possible values in $V_r$ \\	
$O_r$& Order values (integers) of $\mathbf{a}_r$'s possible values $V_r$ \\
$o(v_{r,g})$& Order value (integer) of possible value $v_{r,g}$ \\
$\mathbf{Q}$ and $C_m$& Object-cluster affiliation matrix and $m$-th cluster \\	
$q_{i,m}$& $(i,m)$-th entry of $\mathbf{Q}$ indicating affiliation of $\mathbf{x}_i$ to $C_m$\\
$\Theta(\mathbf{x}_i,C_m;O)$& Overall order distance between $\mathbf{x}_i$ and $C_m$\\
$\theta(x_{i,r},\textbf{p}_{m,r})$& Order distance between $\mathbf{x}_i$ and $C_m$ reflected by $\mathbf{a}_r$\\
$\mathbf{D}$ and $\mathbf{d}_{i,r}$& Order distance matrix and its $(i,r)$-th entry\\
$d_{i,r,g}$& Order distance between $x_{i,r}$ and $v_{r,g}$, also the $g$-th value of $\mathbf{d}_{i,r}$\\
$\mathbf{P}$ and $\mathbf{p}_{m,r}$& Conditional probability matrix and its $(m,r)$-th entry\\
$p_{m,r,g}$& Occurrence probability of $v_{r,g}$ in $C_m$, also the $g$-th value of $\mathbf{p}_{m,r}$\\
$O_{r,m}$& Order values of $\mathbf{a}_r$'s possible values $V_r$ obtained through $C_m$\\
$L_{m,r}$& A fraction of $L$ jointly contributed by $C_m$ and $\mathbf{a}_r$\\
\bottomrule
\end{tabular}}
\end{table}

\rtwo{It is noteworthy that fixing $\text{dist}(v_{r,g},v_{r,h})\equiv1$ by the Hamming distance can limit the possibility of obtaining a context-aware distance metric, while most categorical data distance metrics based on statistics do not reveal a value order to ensure compatibility with the naturally ordered values of a numerical attribute under the Euclidean distance metric. This paper, therefore, aims to learn a set of optimal orders $O^*=\{O^*_1, O^*_2,..., O^*_s\}$ corresponding to the $s^{\{c\}}$ attributes in $A^{\{c\}}$, so that the order-guided distance metric is compatible with numerical attributes and is adaptive to clustering tasks.}
For crisp clustering that partitions $X$ into $k$ non-overlapping sample sets $C=\{C_1, C_2,..., C_k\}$, the objective with our order distance learning can be formalized as the problem of minimizing: 
\begin{equation}\label{eq:obj_pre}
	L(\mathbf{Q},O)=\sum_{m=1}^k\sum_{i=1}^nq_{i,m}\cdot\Theta(\mathbf{x}_i,C_m;O)
\end{equation}
where $\mathbf{Q}$ is an $n\times k$ matrix with its $(i,m)$-th entry $q_{i,m}$ indicating the affiliation between sample $\mathbf{x}_i$ and cluster $C_m$. Specifically, $q_{i,m}=1$ indicates that $\mathbf{x}_i$ belongs to $C_m$, while $q_{i,m}=0$ indicates that $\mathbf{x}_i$ belongs to a cluster other than $C_m$, which can be written as:
\begin{equation}
	\label{eq:qim_pre}
	q_{i,m}=\left\{
	\begin{array}{ll}
		1,  & \text{if}\ m=\arg\min\limits_y\Theta(\mathbf{x}_i,C_y;O)\\
		0,  & \text{otherwise}\\
	\end{array}\right.
\end{equation}
with $\sum_{i=\rthree{1}}^kq_{i,m}=1$. The distance $\Theta(\mathbf{x}_i, C_m; O)$ reflects the dissimilarity between $\mathbf{x}_i$ and $C_m$ computed based on order relationship $O$. \rthree{Since the Euclidean distance of numerical attributes $A^{\{u\}}$ is known, the problem to be solved by this paper can be specified as how to jointly learn $O$ of categorical attributes $A^{\{c\}}$} and partition $\mathbf{Q}$ to minimize $L$. 
Frequently used symbols are sorted out in Table~\textcolor{blue}{\ref{tb:symbol}}.

\subsection{Order Distance}

\rtwo{Order distance is the key component of the clustering objective in Eq.~(\textcolor{blue}{\ref{eq:obj_pre}}), as it acts to decide the sample-cluster affiliation for data partitioning, indicates whether the current order is conducive to approaching the minimization of the clustering objective, and provides a quantitative basis for order learning. Given the order $O$ representing the attribute value relationship, the sample-cluster distance $\Theta(\mathbf{x}_i, C_m; O)$ can be expressed as:}

\begin{equation}
	\label{eq:dist_m}
	\Theta(\mathbf{x}_i,C_m;O)=\frac{1}{s}\sum_{r=1}^{s}\theta(x_{i,r},\textbf{p}_{m,r})
\end{equation}
\rtwo{where $\theta(x_{i,r},\textbf{p}_{m,r})$ is the order distance between $\mathbf{x}_i$ and $C_m$ reflected by the $r$-th attribute $\mathbf{a}_r$, indicating how dissimilar is the sample value $x_{i,r}$ to $\mathbf{a}_r$'s values in $C_m$. To sufficiently exploit the statistics of $C_m$, probability distribution $\mathbf{p}_{m,r}$ of $\mathbf{a}_r$'s possible values $V_r$ within $C_m$ is considered to derive $\theta(x_{i,r},\textbf{p}_{m,r})$:}
\begin{equation}
	\label{eq:dist_mr}
	\theta(x_{i,r},\textbf{p}_{m,r})=\mathbf{d}_{i,r}^\top\mathbf{p}_{m,r}
\end{equation}
where $\mathbf{p}_{m,r}$ is an $l_r$-dimensional vector $[p_{m,r,1},p_{m,r,2},...,p_{m,r,l_r}]$ describing the probability distribution of $V_r$ within cluster $C_m$, with its $g$-th value $p_{m,r,g}$ indicating the occurrence probability of $v_{r,g}$ in cluster $C_m$ obtained by:
\begin{equation}
	\label{eq:occ_prob}
	p_{m,r,g}=\frac{card(X_{r,g}\cap C_m)}{card(C_m)}.
\end{equation}
Here, $X_{r,g}=\{\mathbf{x}_i|x_{i,r}=v_{r,g}\}$ is collection of all the samples with their $r$-th values equal to $v_{r,g}$ in $X$, and the cardinality function $card(\cdot)$ counts the number of elements in a set.
$\mathbf{d}_{i,r}$ in Eq.~(\textcolor{blue}{\ref{eq:dist_mr}}) is the order differences between $x_{i,r}$ and all the values in $V_r$, which is also expressed as a vector $[d_{i,r,1},d_{i,r,2},...,d_{i,r,l_r}]$. Its $g$-th value can be computed by:
\begin{equation}
	\label{eq:ord_dist}
	d_{i,r,g}=\frac{|o(x_{i,r})-o(v_{r,g})|}{l_r-1}
\end{equation}
where $l_r-1$ normalizes the differences into $[0,1]$ to ensure comparability across different attributes. \rtwo{Eq.~(\textcolor{blue}{\ref{eq:ord_dist}}) connects the value order to the objective function, and thus the order distance defined in Eq.~(\textcolor{blue}{\ref{eq:dist_mr}}) can quantify the appropriateness of the current order $O_r$ of $\mathbf{a}_r$. That is, $O_r$ corresponding to smaller $\theta(x_{i,r},\textbf{p}_{m,r})$ with $\mathbf{x}_i\in C_m$ is preferable as it indicates a more compact $C_m$ and a higher contribution in minimizing $L$. Subsequently, Section~\textcolor{blue}{\ref{sct: estimation}} introduces the order learning based on order distance.}

\subsection{Order Learning}\label{sct: estimation}

\rtwo{Given data partition $\mathbf{Q}$,} the optimal order $O^*$ can be expressed as:
\begin{equation}
	\label{eq:optimal_order}
	O^*=\arg\min_{O}L(\mathbf{Q},O).
\end{equation}
A simple but laborious solution is to search different order combinations across all the attributes. However, there are a total of $\prod_{r=1}^dl_r!$ combinations, and each combination should be evaluated by considering all $n$ samples through the clustering objective function, which makes the search almost infeasible for large-scale data. A relaxed efficient solution is to search the optimal order within clusters by assuming independence of attributes. Accordingly, optimal order of $\mathbf{a}_r$ w.r.t. $C_m$ denoted as $O^*_{r,m}=\{o^*_m(v_{r,1}),o^*_m(v_{r,2}),...,o^*_m(v_{r,l_r})\}$ can be obtained through within-cluster traversal search:
\begin{equation}\label{eq:relax_order}
        O^*_{r,m}=\arg\min_{O_r}L_{m,r}
\end{equation}
where $L_{m,r}$ is an inner sum component of the objective $L$. More specifically, $L_{m,r}$ is jointly contributed by the $r$-th attribute and $m$-th cluster, which can be derived from Eqs.~(\textcolor{blue}{\ref{eq:obj_pre}}) and~(\textcolor{blue}{\ref{eq:dist_m}}) as:
\begin{equation}\label{eq:obj_component}
	L_{m,r}=\sum_{\mathbf{x}_i\in C_m}\theta(x_{i,r},\textbf{p}_{m,r}).
\end{equation}


\rtwo{However, the traversal search for the optimal order is still with an $O(\varsigma!)$ time complexity in the worst case where $\varsigma=\max(l_1,l_2,...,l_d)$, which can still be laborious for attributes with a large number of possible values $\varsigma$. Fortunately, such an optimal order searching problem is a simple special case of the well-known Quadratic Assignment Problem (QAP)~\textcolor{blue}{\cite{QAP_Survey, QAP_PAMI1,QAP_TSYST}}, for which an efficient Optimal Linear Ordering (OLO)~\textcolor{blue}{\cite{OLO}} algorithm with time complexity $O(\varsigma\log\varsigma)$ has been proposed with optimal solution guarantee. OLO solves a circuit routing cost optimization. It assumes that there are a set of $\varsigma$ pins with $\varsigma(\varsigma-1)/2$ wire connections, and there are $\varsigma$ holes in a line with adjacent holes at unit distances apart to put the $\varsigma$ pins. If the number of wires connecting the pins is given, OLO can arrange the pins into holes to minimize the overall wire length to save cost. Intuitively, the solution follows a unimodal arrangement, where the pins that are more densely connected with the other pins (i.e., with more wire linkages) are placed in the middle holes. Corresponding to our problem, how to define the number of wires between the pins, i.e., the link density between possible values within the same cluster, is the prerequisite to implement OLO.}

\begin{figure}[!t]	
\centerline{\includegraphics[width=3.4in]{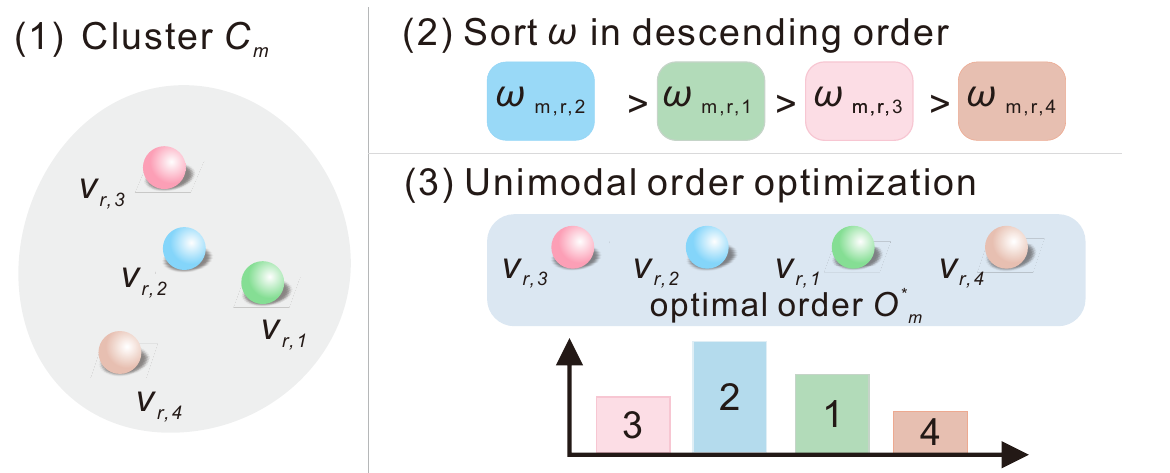}}
\caption{\rtwo{Illustration of the order learning process. (1) Four values of $\mathbf{a}_r$ appear in $C_m$; (2) Their link density $\omega_{m,r,u}$ are quantified and ranked; (3) Order optimization by Eq.~(\textcolor{blue}{\ref{eq:compute_value_order}}).}}
\label{fig:OrderLearning}	
\end{figure}

\rtwo{Using the occurrence probability $p_{m,r,g}$ of a possible value within a cluster $C_m$ defined in Eq.~(\textcolor{blue}{\ref{eq:occ_prob}}) as the link density is a straightforward possible way. This is equivalent to quantifying the number of wire linkages by the occurrence frequency of each possible value, and the rationality lies in that the values with a larger proportion have a stronger dominance over the objective function. However, taking into account only the within-cluster probabilities of the values makes the OLO process relatively isolated, i.e., less connected with the clustering objective in the alternating optimization process, potentially hindering the convergence of the algorithm. Thus, contribution of a possible value $v_{r,g}$ to a cluster $C_m$ in forming the objective function is also derived as:
\begin{equation}\label{eq:obj_sub_component}
	L_{m,r,g}=\sum_{x_{i,r}\in \{X_{r,g}\cap C_m\}}\theta(x_{i,r},\textbf{p}_{m,r}),
\end{equation}
which is denoted as the form of a component of $L_{m,r}$ satisfying:
\begin{equation}\label{eq:obj_component_relation}
    L_{m,r}=\sum_{g=1}^{l_r} L_{m,r,g}.
\end{equation}
By simultaneously considering $p_{m,r,g}$ and $L_{m,r,g}$, the link density of $v_{r,g}$ to the other possible values within $C_m$ can be quantified as:
\begin{equation}
    \label{eq_count_mid_value}
    \omega_{m,r,g}=\frac{p_{m,r,g}}{L_{m,r, g}},  
\end{equation}  
and the rationality of introducing $L_{m,r, g}$ is analyzed as Remark~\textcolor{blue}{\ref{rmk:connect_strength}}.}

\rthree{
\begin{remark} \label{rmk:connect_strength}
\textbf{Rationality of $L_{m,r, g}$ in quantifying $\omega_{m,r,g}$.}
$L_{m,r, g}$ in Eq.~(\textcolor{blue}{\ref{eq_count_mid_value}}) acts as a regularization term to reduce the complexity of the learning process. It is obtained based on the partition of the whole dataset and order distance learned in the previous iteration, and forms a balance with $p_{m,r,g}$ that only considers the value distribution within the current $C_m$. Intuitively, a smaller $L_{m,r, g}$ indicates that $v_{r,g}$ contributes less to $L_{m,r}$. This results in a larger $\omega_{m,r,g}$, emphasizing its importance in the OLO order optimization process, and further consolidating its role in minimizing the clustering objective function. This effectively circumvents optimizing $L$ in the large order space and improves the stability of the learned order in adjacent iterations.
\end{remark}}

\rtwo{With link density defined by Eq.~(\textcolor{blue}{\ref{eq_count_mid_value}}), the unimodal order arrangement of OLO to minimize linkage cost can be expressed as:
\begin{equation} \label{eq:compute_value_order}
    o^*_{m}(v_{r,g}) = \lceil \frac{l_r}{2}\rceil - (-1)^{\eta+1}( \left\lfloor \frac{\eta}{2} \right\rfloor)
\end{equation}
where $\eta=\text{rank}(\omega_{m,r,g})$ is the rank of $\omega_{m,r,g}$ in the descending order of $\{\omega_{m,r,1}, \omega_{m,r,2}, ..., \omega_{m,r,l_r}\}$, and $\lceil \cdot \rceil$ and $\left\lfloor \cdot \right\rfloor$ take the ceiling and floor of a value, respectively. Such an arrangement put the values with the highest link density to the center, and arrange the remaining values in descending order of their link density, alternating between the left and right sides of the value with the highest link density. The core idea of this process is to place the value with the highest link density in the middle of the order to effectively reduce its overall order distance with other values, thereby minimizing the linkage cost. For more OLO derivation and its $O(\varsigma\log\varsigma)$ time complexity analysis, please refer to~\textcolor{blue}{\cite{OLO}}.}

\rthree{A possible alternative solution is to obtain differentiated orders across different cluster groups. However, each order subtly reflects the relationship between the values of the corresponding attribute. If the order is differentiated w.r.t. different cluster groups, the overly distorted distance metric space derived from the inconsistent orders may make the algorithm prone to falling into local optima or lead to learning divergence.}
By searching for the optimal ranking in each cluster (as shown in Figure~\textcolor{blue}{\ref{fig:OrderLearning}}) and then performing a linear combination based on the cluster size, the overall consensus ranking of each possible value $v_{r,g}$ can be obtained by:
\begin{equation}
	\label{eq:weight_mean}
	{o}(v_{r,g})=\sum_{m=1}^k\left(\frac{card(C_m)}{n}\cdot o^*_m(v_{r,g})\right).
\end{equation}
Since the value of ${o}(v_{r,g})$ is not necessarily an integer, the obtained rankings $\{{o}(v_{r,1}),{o}(v_{r,2}),...,{o}(v_{r,l_r})\}$ of an attribute is sorted in ascending order to obtain the final ranking of $V_r$, and the corresponding integer order $\tilde{O}_r = \{\tilde{o}(v_{r,1}), \tilde{o}(v_{r,2}), ..., \tilde{o}(v_{r,l_r})\}$ is obtained. Consequently, all orders of $A^{\{c\}}$ are denoted as $\tilde{O}=\{\tilde{O}_1,\tilde{O}_2,...,\tilde{O}_s\}$. \rtwo{The optimal order learning described by Eqs.~(\textcolor{blue}{\ref{eq:relax_order}}) - (\textcolor{blue}{\ref{eq:weight_mean}}) considers both the within-cluster statistics $p_{m,r,g}$ and the optimization of the overall objective function $L_{m,r,g}$ as discussed in Remark~\textcolor{blue}{\ref{rmk:connect_strength}}, ensuring the consistency of the objectives of order learning and clustering. $\tilde{O}$ is utilized to form distance metric by Eqs.~(\textcolor{blue}{\ref{eq:dist_m}}) and (\textcolor{blue}{\ref{eq:dist_mr}}) for clustering in the alternating optimization process presented in Section~\textcolor{blue}{\ref{sct: optimization}}.}

\subsection{Joint Order and Cluster Learning}\label{sct: optimization}
By integrating the order learning with clustering, the objective can be rewritten in a more detailed form based on Eqs.~(\textcolor{blue}{\ref{eq:obj_pre}}) - (\textcolor{blue}{\ref{eq:dist_mr}}) as:
\begin{equation}\label{eq:obj}
	L(\mathbf{Q},O)=\sum_{m=1}^k\sum_{i=1}^nq_{i,m}\cdot\left(\frac{1}{s^{\{c\}}}\sum_{\textbf{a}_r\in\textbf{A}^{\{c\}}}\mathbf{d}_{i,r}^\top\mathbf{p}_{m,r}\right)
\end{equation}
where the order distance $\mathbf{D}=\{\mathbf{d}_{i,r}|i\in\{1,2,...,n\}, \textbf{a}_r\in\textbf{A}^{\{c\}}\}$ is dependent to the order $O$ according to Eqs.~(\textcolor{blue}{\ref{eq:ord_dist}}) and~(\textcolor{blue}{\ref{eq:compute_value_order}}), and the probability distribution $\mathbf{P}=\{\mathbf{p}_{m,r}|m=\{1,2...,k\}, r\in\{1,2,...,s^{\{c\}}\}\}$ is dependent to the partition $\mathbf{Q}$ according to Eqs.~(\textcolor{blue}{\ref{eq:qim_pre}}) and~(\textcolor{blue}{\ref{eq:occ_prob}}). Typically, $\mathbf{Q}$ and $O$ are iteratively computed to minimize $L$. Specifically, given $\hat{O}$, $\mathbf{D}$ is updated accordingly, then a new $\mathbf{Q}$ is computed by:
\begin{equation}
	\label{eq:qim}
	q_{i,m}=\left\{
	\begin{array}{ll}
		1,  & \text{if}\ m=\arg\min\limits_y\frac{1}{s^{\{c\}}}\sum_{r=1}^{s^{\{c\}}}\mathbf{d}_{i,r}^\top\mathbf{p}_{y,r}\\
		0,  & \text{otherwise}\\
	\end{array}\right.
\end{equation}
with $i=\{1,2,...,n\}$ and $m=\{1,2,...,k\}$. Eq.~(\textcolor{blue}{\ref{eq:qim}}) is a detailed version of Eq.~(\textcolor{blue}{\ref{eq:qim_pre}}) that adopts the order distance defined by Eqs.~(\textcolor{blue}{\ref{eq:dist_m}}) and~(\textcolor{blue}{\ref{eq:dist_mr}}). Then, $\mathbf{P}$ is updated according to the new $\mathbf{Q}$. When the iterative updating of $\mathbf{Q}$ and $\mathbf{P}$ converges, we obtain $O$ according to Eqs.~(\textcolor{blue}{\ref{eq:relax_order}}) - (\textcolor{blue}{\ref{eq:weight_mean}}), and update $\mathbf{D}$ accordingly. \rtwo{To ensure effective initialization of the optimization process, $k$-modes is implemented once and the clustering results are utilized as the starting point, which has been widely adopted in clustering optimization.} 
In summary, $L$ is minimized by iteratively solving the following two sub-problems: 1) fix $\hat{\mathbf{Q}}$ and $\hat{\mathbf{P}}$, solve the minimization problem $L(\hat{\mathbf{Q}},O)$ by searching $O$ and updating $\mathbf{D}$ accordingly, and 2) fix $\hat{O}$ and $\hat{\mathbf{D}}$, solve the minimization problem $L(\mathbf{Q},\hat{O})$ by iteratively computing $\mathbf{Q}$, and updating $\mathbf{P}$ according to $\mathbf{Q}$, until convergence.  
The whole algorithm named Order and Cluster Learning (OCL) is summarized as Algorithm~\textcolor{blue}{\ref{alg:OCL}}.

\rthree{Since the learned order distance of each attribute resides in a one-dimensional metric space compatible with the attribute-level Euclidean distance, it can be seamlessly integrated with numerical attributes for homogeneous computation. This enables straightforward extension of OCL to mixed data clustering. Specifically, the order distance metric of the categorical attributes is first learned using OCL, and then combined with numerical attributes using a conventional clustering algorithm such as $k$-means~\textcolor{blue}{\cite{kms}}. Crucially, this decouples order distance learning of categorical attributes from the deterministic Euclidean distance of numerical attributes. Furthermore, the compatibility of learned ordinal metric and the Euclidean distance space considerably relieves information loss during joint clustering, enhancing performance on mixed datasets.}

\begin{algorithm}[!t]
\caption{OCL: Order and Cluster Learning Algorithm}
\begin{algorithmic}[1]
\label{alg:OCL}
\REQUIRE Dataset $X$, number of \rtwo{sought} clusters $k$
\ENSURE Partition $\mathbf{Q}$, order $O$
\STATE Initialization: Set outer loop counter and convergence mark by $E\leftarrow 0$ and $Conv\_O\leftarrow False$, respectively; Randomly partition $X$ into $k$ clusters $C$, obtain $\mathbf{Q}^{\{I\}}$ and $\mathbf{P}^{\{I\}}$ accordingly; Randomly initialize $O^{\{E\}}$, update $\mathbf{D}^{\{E\}}$ accordingly;
\WHILE{$Conv\_O=False$}
\STATE $E\leftarrow E+1$; 
Compute $O^{\{E\}}$ by Eqs.~(\textcolor{blue}{\ref{eq:relax_order}}) - (\textcolor{blue}{\ref{eq:weight_mean}}); Update $\mathbf{D}^{\{E\}}$ accordingly; Set inner loop counter and convergence mark by $I\leftarrow 0$ and $Conv\_I\leftarrow False$, respectively;
\WHILE{$Conv\_I=False$}
\STATE $I\leftarrow I+1$; Compute $\mathbf{Q}^{\{I\}}$ by Eq.~(\textcolor{blue}{\ref{eq:qim}}); Update $\mathbf{P}^{\{I\}}$ accordingly; Obtain $L(\mathbf{Q}^{\{I\}},O^{\{E\}})$;
\IF {$L(\mathbf{Q}^{\{I\}},O^{\{E\}})\geq L(\mathbf{Q}^{\{I-1\}},O^{\{E\}})$}
\STATE $Conv\_I\leftarrow True$; $\mathbf{Q}^{\{E\}}=\mathbf{Q}^{\{I\}}$;
\ENDIF
\ENDWHILE
\IF {$L(\mathbf{Q}^{\{E\}},O^{\{E\}})\geq L(\mathbf{Q}^{\{E-1\}},O^{\{E-1\}})$}
\STATE $Conv\_O\leftarrow True$;
\ENDIF
\ENDWHILE
	\end{algorithmic}
\end{algorithm}

\subsection{Complexity and Convergence Analysis}\label{sct:theoretical_analysis}

The \rone{time and space complexity} and convergence of OCL are analyzed below. 

\begin{theorem} \label{the:time_analysis}
\rone{The time complexity of OCL is $O(EInks\varsigma+Enks\varsigma\log\varsigma)$.}
\end{theorem}
\textit{Proof.} 
Assume it involves $I$ inner iterations (lines 4 - 9 in Algorithm~\textcolor{blue}{\ref{alg:OCL}}) to compute $\mathbf{Q}$ and $\mathbf{P}$, and $E$ outer iterations (lines 2 - 13 in Algorithm~\textcolor{blue}{\ref{alg:OCL}}) to search $O$ and update $\mathbf{D}$. For worst-case analysis, $\varsigma=\max(l_1,l_2,...,l_s)$ is adopted to indicate the maximum number of possible values of a dataset.

For each of the $I$ inner iterations, $\mathbf{P}$ is obtained upon the $n$ data samples with complexity $O(nks\varsigma)$, and $\mathbf{D}$ is obtained upon the $\varsigma$ values of $s$ attributes on $n$ samples with complexity $O(nd\varsigma)$. Then the $n$ samples are partitioned to $k$ clusters with complexity $O(nks\varsigma)$. For $I$ iterations in total, the complexity is $O(Inks\varsigma)$.

\rone{For each of the $E$ outer iterations, a new order of $\varsigma$ possible values of each attribute within each cluster is searched by OLO algorithm described by Eqs.~(\textcolor{blue}{\ref{eq:relax_order}}) - (\textcolor{blue}{\ref{eq:weight_mean}}), and the complexity for $k$ clusters and $s$ attributes is thus $O(nks\varsigma\log\varsigma)$. }

\rone{For $E$ outer iterations with considering the $I$ inner iterations, the overall time complexity of OCL is $O(EInks\varsigma+Enks\varsigma\log\varsigma)$.}\hfill\Square

\begin{theorem} \label{the:space_analysis}
\rone{The space complexity of OCL is $O(ns + nk + ns\varsigma + ks\varsigma)$.}
\end{theorem}

\rone{\textit{Proof.}
OCL requires $O(ns)$ space for the dataset $X$, $O(nk)$ for the cluster assignment matrix $\mathbf{Q}$, $O(ns\varsigma)$ for the order distance matrix $\mathbf{D}$, and $O(ks\varsigma)$ space each for the conditional probability matrix $\mathbf{P}$ and the link density.  }\hfill\Square

\begin{remark}\label{rmk:time}
\textbf{Scalability of OCL.} The essential characteristic of categorical data is that the number of possible values $\varsigma$ for an attribute \rtwo{is finite satisfying $\varsigma<n$ for real benchmark datasets as shown in Table~\textcolor{blue}{\ref{tb:detaile_infor}}. Therefore, the time and space complexity of OCL scales linearly with $n$, $k$, and $s$, and are also relatively insensitive to $\varsigma$, indicating high scalability of the proposed OCL algorithm.}
\end{remark}





\begin{table*}[!t]
\caption{Detailed statistics of the 20 datasets. ``$s^{\{n\}}$'', ``$s^{\{o\}}$'', and ``$s^{\{u\}}$'' indicate the numbers of nominal, ordinal, and numerical attributes, respectively. ``$l_r$'' indicates the number of possible values of categorical (ordinal + nominal) attributes. $\vartheta$, $\varsigma$, and $\varepsilon$ indicate the averaged, maximum, and minimum numbers of possible values for the categorical attributes of each dataset, respectively. ``$n$'' is the number of samples. $k^*$ is the true number of clusters.}
\label{tb:detaile_infor}
\centering
\resizebox{2.0\columnwidth}{!}{
\begin{tabular}{lcc|ccc|cccc|cc}
\toprule
no.&Data &abbrev.& $s^{\{n\}}$&$s^{\{o\}}$&$s^{\{u\}}$ & $l_r$ & $\vartheta$ &$\varsigma$ &$\varepsilon$& $n$ & $k^*$\\
\midrule
1&Soybeans & SB	&35 &0 &0&	$7 \ 2 \  3 \ 3 \ 2  \ 4  \  4 \ 2 \  2 \ 3  \ 2 \ 2  \  4 \  4  \ 2  \ 2 \  2 \ 2 \ 2 \  2\ 2 $	&2.76 &	7&	2 &47 &4\\
2&Nursery School & NS	&1 &7	&0     &$3 \ 5 \  4 \  4 \  3  \  2  \ 3  \ 3$	&3.38	&5	&2 &12960 &4\\
3&Amphibians& AP	&6 &6	&2     &$8  \ 5  \ 8 \ 7 \  8 \ 3  \ 5 \  6 \ 6 \ 6 \  3 \ 2$	&5.58	&8	&2&189 &2\\
4&Inflammations Diagnosis & DS	&0 &5 & 1 	     &$2 \ 2 \ 2 \  2 \ 2	$ &2.00	&2	&2 &120 &2\\
5&Caesarian Section &CS	&1 &3 &0	     &$4 \ 3 \  3 \ 2	$ &3.00	&4	&2 &80 &  2\\
6&Hayes-Roth & HR	&2 &2	&0     &$3   \  4   \  4  \  4$&3.75	&4	&3 &132 &3\\
7&Zoo & ZO	&15 &0	&0     & $2  \   2   \  2  \   2  \  2   \  2 \    2   \  2  \   2   \  2  \   2  \   2  \   6  \   2  \   2 \  2$	& 2.25	&6	&2 &101 &7\\
8&Breast Cancer &BC	&5 & 4 &0	     &$3   \  3   \  2   \  6   \  2   \  6  \  11  \   7  \   3$	&4.78	&11	&2 &286 &2\\
9&Lym & LG	&15&3	&0     &$3   \  4   \  8  \   4   \  2  \   2  \   2   \  2  \   2  \   2  \   2  \   3  \   4   \  4  \   8   \  3   \  2  \   2$	&3.28	&8	&2 &148 &4\\
10&Tic-Tae-Toc & TT	&9 &0 &0	     &$3  \   3  \   3   \  3    \ 3   \  3  \   3  \   3  \   3$	&3.00	&3	&3 &958 &2\\
11&Australia Credit & AC	&0 &8	&6     &2  \   2  \   2  \   2  \   3  \  14    \ 8   \  3 & 4.50	& 14	&2 &690 &2\\
12&Congressional Voting & VT	&16	&0 &0     &$3  \   3    \ 3  \   3   \  3  \   3   \  3   \  3  \   3  \   3  \   3  \   3   \  3  \   3   \  3  \   3$	&3.00	&3	&3 &435 &2\\
13&Connect-4 & CT4  & 42 & 0 &0 & -&3.00 &3 &3 & 67557 & 3\\
14&Obesity Levels & OB & 6 & 2&0& 2  \   2  \   2  \   4  \   2   \  2  \   4  \   5 &2.88 &5 &2  & 2111 & 7\\
15&Auction Verifications & AV & 0 & 6&0& 3 \    4  \   2    \ 2  \   6 \    5 & 3.88& 6&2 & 2043 & 2\\
16&Bank Marking & BM & 6 & 3 &0 &12  \   3 \    4   \  2   \  2   \  2  \   3  \  12 \    4 &4.89 & 12&2 & 45211 & 2\\
17&Chess & CC & 36 & 0 &0& - & 2.02 &3 & 2& 3195 & 2\\
18&Coupon & CR & 7 & 4&0 & 3 \    4  \   3  \   5  \   5   \  2  \   2   \  5 \    6 \   25   \  9 & 6.27 &25 &2 &  12684 & 2\\
\rone{19}& \rone{Heart Failure} & \rone{HF} & \rone{6} & \rone{0} & \rone{8} & \rone{2  \   2 \    2  \   2 \    2  \  47 \  208  \  17  \ 176  \  40  \  27 \  148}  & \rone{56.08} &  \rone{208}& \rone{2} & \rone{299} & \rone{2}\\
\rone{20}& \rone{Multi-attribute-category AC} & \rone{MA}	& \rone{0} & \rone{8}	& \rone{6}     & \rone{2  \   2  \   2  \   2  \   3  \  14    \ 8   \  3  \ 350  \ 215 \  132   \ 23  \ 171  \ 240}	& \rone{83.36}	& \rone{350}	& \rone{2} & \rone{690} & \rone{2}\\

\bottomrule
\end{tabular}}
\end{table*}

\begin{theorem}
OCL algorithm converges to a local minimum in a finite number of iterations.
\end{theorem}
\textit{Proof.} We prove the convergence of the inner and outer loop, i.e., lines 4 - 9 and lines 2 - 13 of Algorithm~\textcolor{blue}{\ref{alg:OCL}}, respectively. 

\begin{lemma}\label{lm:inner}
The inner loop converges to a local minimum of $L(\mathbf{Q},\hat{O})$ in a finite number of iterations with fixed $\hat{O}$. Note that the number of possible partitions $\mathbf{Q}$ is finite for a dataset with a finite number of objects $n$. We then illustrate that each possible partition $\mathbf{Q}$ occurs at most once by OCL. Assume that two optimizers satisfy $\mathbf{Q}^{\{I_1\}}=\mathbf{Q}^{\{I_2\}}$, where $I_1\neq I_2$. Then it is clear that $L(\mathbf{Q}^{\{I_1\}},O^{\{E\}})=L(\mathbf{Q}^{\{I_2\}},O^{\{E\}})$. However, the value sequence of the objective function $\{L(\mathbf{Q}^{\{1\}},O^{\{E\}}), L(\mathbf{Q}^{\{2\}},O^{\{E\}}), ..., L(\mathbf{Q}^{\{I\}},O^{\{E\}})\}$ 
generated by OCL is strictly decreasing. Hence, the result follows.
\end{lemma}

\begin{lemma}\label{lm:outer}
The outer loop converges to a local minimum of $L(\mathbf{Q},O)$ in a finite number of iterations. We note that the status of orders $O$ is in a discrete space and thus the number of possible orders is finite. The number of possible partitions $\mathbf{Q}$ is also finite. Therefore, the number of possible combinations of $\mathbf{Q}$ and $O$ is finite. We then prove that each possible combination of $\mathbf{Q}$ and $O$ appears at most once by OCL. We note that given $\mathbf{Q}^{\{E_1\}}=\mathbf{Q}^{\{E_2\}}$, where $E_1\neq E_2$, the corresponding $O^{\{E_1+1\}}$ and $O^{\{E_2+1\}}$ can be computed accordingly, satisfying $O^{\{E_1+1\}}=O^{\{E_2+1\}}$. Then we can obtain their corresponding partitions $\mathbf{Q}^{\{E_1+1\}}$ and $\mathbf{Q}^{\{E_2+1\}}$ at the convergence of the next inner loop, satisfying $\mathbf{Q}^{\{E_1+1\}}=\mathbf{Q}^{\{E_2+1\}}$. It is clear that $L(\mathbf{Q}^{\{E_1+1\}}, O^{\{E_1+1\}})=L(\mathbf{Q}^{\{E_2+1\}}, O^{\{E_2+1\}})$. However, the sequence $\{L(\mathbf{Q}^{\{1\}}, O^{\{1\}}), L(\mathbf{Q}^{\{2\}}, O^{\{2\}}), ..., L(\mathbf{Q}^{\{E\}}, O^{\{E\}})\}$ is monotonically decreasing. Hence, the result follows.

\end{lemma}

Since both the inner loop and outer loop of Algorithm~\textcolor{blue}{\ref{alg:OCL}} converge to a local minimum, convergence of OCL can be confirmed.
\hfill\Square

\begin{table}[!t]
\caption{\rtwo{Characteristics of compared clustering approaches in terms of the utilization/consideration of Hamming Distance (HD), Statistical Information (SI), Distance Metric Learning (DML), Semantic Order (SO), and Order Learning (OL).}}
\label{tb:method_statistics}
\centering
\resizebox{0.9\columnwidth}{!}{
\begin{tabular}{l|lcccccc}
\toprule
Type &Counterpart& Year & HD & SI & DML &SO & OL\\
\midrule
\multirow{3}{*}{Traditional} & KMD \textcolor{blue}{\cite{kmd}} & 1998 & $\checkmark $&   & &  & \\
& LSM \textcolor{blue}{\cite{lsm}} & 1998 & & $\checkmark $  & & $\checkmark $ & \\
& CBDM \textcolor{blue}{\cite{cbdm_journal}} &2012& &   $\checkmark $ & &  &  \\
\midrule
\multirow{2}{*}{Statistical}& JDM \textcolor{blue}{\cite{jdm}} & 2016 & & $\checkmark $  & & & \\
& UDMC \textcolor{blue}{\cite{udm}} & 2020 & & $\checkmark $ & & $\checkmark $ &  \\
\midrule
\multirow{6}{*}{\makecell{Learning \\ based \\ (SOTA)}}& DLC \textcolor{blue}{\cite{dlc}} & 2020 & &  $\checkmark $  & $\checkmark $ & $\checkmark $  &  \\
 & HDC \textcolor{blue}{\cite{HDC}} & 2022 & &  $\checkmark $  & $\checkmark $ & $\checkmark $  &  \\
& ADC \textcolor{blue}{\cite{ADC}} & 2023 & &  $\checkmark $  & $\checkmark $ & $\checkmark $  &  \\
& MCDC \textcolor{blue}{\cite{ICDCSCai}} & 2024 & &  $\checkmark $  & $\checkmark $ &   &  \\
& AMPHM \textcolor{blue}{\cite{CAIS}} & 2025 & &  $\checkmark $  & $\checkmark $ &   &  \\
& HARR \textcolor{blue}{\cite{HARR}} & 2025 & &  $\checkmark $  & $\checkmark $ & $\checkmark $  &  \\
\midrule
Ours & OCL& $-$ & &  $\checkmark $  & $\checkmark $ &   & $\checkmark $ \\
\bottomrule
\end{tabular}}
\end{table}

\section{Experiments}

A series of experiments have been conducted. Experimental settings and the obtained results are introduced as follows.

\subsection{Experimental Settings}\label{sct:exp_Set}

\textbf{Six experiments} have been conducted: \textbf{(1) Clustering performance evaluation with significance tests (Section~\textcolor{blue}{\ref{sct:CPE}}):} OCL versus conventional/SOTA methods across diverse datasets, with statistical significance analysis; \textbf{(2) Ablation studies (Section~\textcolor{blue}{\ref{sct:ablation}}):} Evaluating OCL's core components and learned order impact through algorithm- and data-aspect ablation.
\textbf{(3) Convergence and efficiency evaluation (Section~\textcolor{blue}{\ref{sct:converge_time}}):} Objective function trajectory recording for convergence analysis and efficiency assessed under scaling sizes of samples $n$ and attributes $s$, \mrone{and number of clusters $k$}. \textbf{(4) Visualization of clustering effect (Section~\textcolor{blue}{\ref{sct:v_c_study}}):} t-SNE embeddings generated via OCL's order distance versus baselines for visual comparison. \textbf{(5) Clustering performance on mixed datasets (Section~\textcolor{blue}{\ref{AP:CPMD}}):} OCL-enhanced methods evaluated on numerical-categorical mixed data, demonstrating effectiveness of the orders learned by OCL. \textbf{(6) Case study on Hayes-Roth (HR) dataset (Section~\textcolor{blue}{\ref{AP: CS}}):} Validation of meaningfulness and human cognition-aligned semantics of the learned orders through analysis and intuitive demonstration of the clustering results on a real dataset. All experiments are coded with MATLAB R2022a and implemented on a workstation with Intel i7-14650 CPU @ 2.20 GHz, 16GB RAM.

\begin{table*}[!t]
\caption{Clustering performance evaluated by CA (with range $[0,1]$). ``$\overline{\textit{AR}}$'' row reports the average performance rankings.}
\label{tb:clustering_ca}
\centering
\resizebox{2.06\columnwidth}{!}{
\begin{tabular}{c|ccccccccccccc}
\toprule
Datasets & KMD & LSM & JDM & CBDM & UDMC & DLC & HDC & ADC & MCDC & AMPHM & HARR & OCL\\
\midrule
SB & 0.8468$\pm$0.15 & 0.8106$\pm$0.16 & 0.7894$\pm$0.14 & 0.7957$\pm$0.15 & 0.7915$\pm$0.18 & 0.8000$\pm$0.14 & 0.7915$\pm$0.15 & 0.7660$\pm$0.13 & 0.5851$\pm$0.12 & 0.7021$\pm$0.00 & \underline{0.9234$\pm$0.12} & \textbf{0.9830$\pm$0.05} \\
NS & 0.3208$\pm$0.05 & 0.2975$\pm$0.04 & 0.2980$\pm$0.04 & $-$ & 0.3074$\pm$0.05 & \underline{0.3386$\pm$0.07} & 0.3208$\pm$0.05 & 0.3208$\pm$0.05 & 0.3370$\pm$0.01 & $-$ & 0.3113$\pm$0.05 & \textbf{0.3573$\pm$0.05} \\
AP & 0.5392$\pm$0.03 & 0.5656$\pm$0.04 & \underline{0.6159$\pm$0.04} & 0.5926$\pm$0.03 & 0.5698$\pm$0.03 & 0.6032$\pm$0.00 & 0.5608$\pm$0.04 & 0.5529$\pm$0.04 & 0.5566$\pm$0.03 & 0.5026$\pm$0.00 & 0.5603$\pm$0.03 & \textbf{0.6222$\pm$0.00} \\
DS & 0.6525$\pm$0.08 & 0.6525$\pm$0.08 & \underline{0.7208$\pm$0.12} & \underline{0.7208$\pm$0.12} & 0.6867$\pm$0.12 & 0.6650$\pm$0.12 & 0.6867$\pm$0.12 & 0.6950$\pm$0.12 & 0.7242$\pm$0.08 & 0.6583$\pm$0.00 & 0.6175$\pm$0.04 & \textbf{0.7458$\pm$0.08} \\
CS & 0.5462$\pm$0.03 & 0.5488$\pm$0.06 & 0.5788$\pm$0.07 & 0.5875$\pm$0.04 & 0.5825$\pm$0.05 & \textbf{0.6350$\pm$0.03} & 0.5837$\pm$0.05 & 0.5525$\pm$0.04 & 0.5863$\pm$0.04 & 0.5000$\pm$0.00 & 0.6075$\pm$0.02 & \underline{0.6313$\pm$0.04} \\
HR & 0.3758$\pm$0.02 & 0.3886$\pm$0.03 & 0.3917$\pm$0.02 & 0.3924$\pm$0.04 & 0.3773$\pm$0.02 & 0.3697$\pm$0.04 & 0.4000$\pm$0.03 & 0.3758$\pm$0.02 & 0.3939$\pm$0.03 & \underline{0.4167$\pm$0.00} & 0.3333$\pm$0.00 & \textbf{0.4326$\pm$0.02} \\
ZO & 0.6624$\pm$0.12 & 0.6733$\pm$0.11 & 0.6743$\pm$0.11 & 0.6752$\pm$0.10 & 0.6693$\pm$0.10 & 0.6802$\pm$0.08 & 0.6733$\pm$0.10 & 0.6743$\pm$0.11 & 0.6228$\pm$0.07 & 0.6832$\pm$0.00 & \underline{0.7059$\pm$0.05} & \textbf{0.7792$\pm$0.10} \\
BC & 0.5472$\pm$0.04 & 0.5594$\pm$0.07 & 0.6392$\pm$0.09 & 0.6346$\pm$0.09 & 0.5413$\pm$0.06 & 0.6140$\pm$0.09 & 0.6052$\pm$0.10 & 0.6185$\pm$0.10 & 0.5804$\pm$0.08 & \underline{0.6399$\pm$0.00} & 0.5266$\pm$0.06 & \textbf{0.6650$\pm$0.05} \\
LG & 0.4277$\pm$0.06 & 0.4446$\pm$0.04 & 0.4547$\pm$0.06 & 0.4905$\pm$0.06 & 0.4466$\pm$0.05 & 0.5095$\pm$0.08 & 0.4959$\pm$0.05 & 0.5101$\pm$0.04 & 0.4743$\pm$0.04 & \underline{0.5308$\pm$0.00} & 0.4959$\pm$0.05 & \textbf{0.5426$\pm$0.02} \\
TT & 0.5569$\pm$0.05 & 0.5593$\pm$0.04 & 0.5563$\pm$0.04 & 0.5408$\pm$0.03 & 0.5386$\pm$0.03 & 0.5637$\pm$0.03 & 0.5465$\pm$0.04 & 0.5471$\pm$0.04 & \textbf{0.6078$\pm$0.03} & 0.5021$\pm$0.00 & 0.5640$\pm$0.04 & \underline{0.5785$\pm$0.03} \\
AC & 0.7281$\pm$0.12 & \underline{0.7783$\pm$0.08} & 0.6335$\pm$0.12 & 0.6354$\pm$0.12 & 0.7683$\pm$0.08 & 0.7488$\pm$0.14 & 0.6762$\pm$0.14 & 0.6293$\pm$0.14 & 0.7717$\pm$0.10 & 0.5565$\pm$0.00 & 0.7703$\pm$0.08 & \textbf{0.8206$\pm$0.08} \\
VT & 0.8625$\pm$0.01 & 0.8699$\pm$0.00 & 0.8699$\pm$0.00 & \underline{0.8761$\pm$0.00} & 0.8667$\pm$0.00 & 0.8531$\pm$0.09 & 0.8736$\pm$0.00 & 0.8713$\pm$0.00 & 0.8667$\pm$0.00 & 0.6092$\pm$0.00 & 0.8736$\pm$0.00 & \textbf{0.8943$\pm$0.00} \\
CT4 & \underline{0.4000$\pm$0.02} & 0.3609$\pm$0.01 & 0.3696$\pm$0.02 & 0.3651$\pm$0.02 & 0.3819$\pm$0.03 & 0.3618$\pm$0.01 & 0.3924$\pm$0.03 & 0.3879$\pm$0.03 & 0.3914$\pm$0.03 & 0.3514$\pm$0.01 & 0.3824$\pm$0.00 & \textbf{0.4121$\pm$0.02} \\
OB & 0.3715$\pm$0.03 & 0.3697$\pm$0.02 & 0.3594$\pm$0.03 & 0.3673$\pm$0.03 & 0.3720$\pm$0.02 & 0.3724$\pm$0.04 & \underline{0.3797$\pm$0.02} & 0.3617$\pm$0.04 & 0.3765$\pm$0.03 & 0.3717$\pm$0.02 & 0.3512$\pm$0.00 & \textbf{0.3935$\pm$0.01} \\
AV & 0.6251$\pm$0.08 & 0.6035$\pm$0.10 & 0.6291$\pm$0.12 & \textbf{0.6695$\pm$0.10} & 0.6065$\pm$0.10 & 0.6150$\pm$0.06 & 0.6441$\pm$0.14 & 0.6364$\pm$0.10 & 0.6325$\pm$0.12 & 0.6114$\pm$0.10 & 0.6125$\pm$0.00 & \underline{0.6637$\pm$0.07} \\
BM & 0.5730$\pm$0.06 & 0.5792$\pm$0.04 & 0.5959$\pm$0.07 & 0.6011$\pm$0.07 & 0.5709$\pm$0.03 & 0.5944$\pm$0.05 & \underline{0.6064$\pm$0.12} & 0.5457$\pm$0.03 & 0.5543$\pm$0.04 & 0.5573$\pm$0.02 & 0.5039$\pm$0.00 & \textbf{0.6127$\pm$0.00} \\
CC & 0.5562$\pm$0.04 & 0.5572$\pm$0.03 & 0.5632$\pm$0.03 & 0.5433$\pm$0.03 & 0.5587$\pm$0.02 & 0.5342$\pm$0.04 & 0.5248$\pm$0.02 & 0.5438$\pm$0.03 & 0.5395$\pm$0.04 & \underline{0.5668$\pm$0.03} & 0.5012$\pm$0.00 & \textbf{0.5726$\pm$0.05} \\
CR & 0.5209$\pm$0.02 & 0.5247$\pm$0.02 & 0.5185$\pm$0.02 & 0.5297$\pm$0.02 & 0.5245$\pm$0.01 & 0.5488$\pm$0.03 & \underline{0.5541$\pm$0.00} & 0.5466$\pm$0.02 & 0.5504$\pm$0.02 & 0.5246$\pm$0.02 & 0.5510$\pm$0.00 & \textbf{0.5590$\pm$0.02} \\
\rone{HF}  & \rone{0.5482$\pm$0.05} & \rone{0.5301$\pm$0.06} & \rone{0.5375$\pm$0.05 }& \rone{0.5284$\pm$0.07} & \rone{0.5385$\pm$0.10} & \rone{0.5478$\pm$0.08 }& \rone{0.5465$\pm$0.04 }& \rone{0.5431$\pm$0.11 }& \rone{0.5411$\pm$0.08 }& \rone{\underline{0.5488$\pm$0.00} }& \rone{$-$} & \rone{ \textbf{0.5652$\pm$0.03}} \\
\rone{MA}  & \rone{0.6961$\pm$0.09 }& \rone{0.7097$\pm$0.03 }& \rone{0.6358$\pm$0.12 }& \rone{\underline{0.7164$\pm$0.07}} & \rone{\underline{0.7164$\pm$0.03}} & \rone{0.6654$\pm$0.10} & \rone{0.7174$\pm$0.06} & \rone{0.7154$\pm$ 0.07} & \rone{0.6649$\pm$0.09} & \rone{0.5594$\pm$0.00} & \rone{-}& \rone{\textbf{0.7333$\pm$0.06}} \\

\midrule
$\overline{	\textit{AR}}$ &  \rone{7.80} &  \rone{7.78} &  \rone{7.12} &  \rone{6.10} &  \rone{7.72} &  \rone{5.95} &  \rone{5.33} &  \rone{7.05 }&  \rone{6.33} &  \rone{8.07} &  \rone{7.60} &  \rone{1.15} \\
\bottomrule
\end{tabular}
}
\end{table*}

\begin{table*}[!t]
\caption{Clustering performance evaluated by ARI (with range $[-1,1]$). ``$\overline{\textit{AR}}$'' row reports the average performance rankings.}
\label{tb:clustering_ar}
\centering
\resizebox{2.06\columnwidth}{!}{
\begin{tabular}{c|ccccccccccccc}
\toprule
Datasets & KMD & LSM & JDM & CBDM & UDMC & DLC & HDC & ADC & MCDC & AMPHM & HARR & OCL\\
\midrule
SB & 0.7638$\pm$0.18 & 0.7555$\pm$0.19 & 0.7232$\pm$0.19 & 0.7407$\pm$0.19 & 0.7307$\pm$0.22 & 0.7148$\pm$0.20 & 0.7350$\pm$0.19 & 0.6968$\pm$0.17 & 0.2736$\pm$0.18 & 0.5949$\pm$0.00 & \underline{0.8650$\pm$0.22} & \textbf{0.9620$\pm$0.12} \\
NS & 0.0376$\pm$0.03 & 0.0341$\pm$0.02 & 0.0283$\pm$0.02 & $-$ & 0.0393$\pm$0.03 & \underline{0.0779$\pm$0.08} & 0.0376$\pm$0.03 & 0.0376$\pm$0.03 & 0.0337$\pm$0.02 & $-$ & 0.0309$\pm$0.03 & \textbf{0.1015$\pm$0.06} \\
AP & 0.0022$\pm$0.01 & 0.0171$\pm$0.02 & \textbf{0.0530$\pm$0.03} & 0.0322$\pm$0.02 & 0.0178$\pm$0.02 & 0.0355$\pm$0.00 & 0.0150$\pm$0.02 & 0.0066$\pm$0.02 & 0.0005$\pm$0.02 & -0.0055$\pm$0.00 & 0.0076$\pm$0.02 & \underline{0.0524$\pm$0.00} \\
DS & 0.1077$\pm$0.13 & 0.1077$\pm$0.13 & \underline{0.2347$\pm$0.20} & \underline{0.2347$\pm$0.20} & 0.1844$\pm$0.20 & 0.1556$\pm$0.19 & 0.1844$\pm$0.20 & 0.1921$\pm$0.19 & 0.2108$\pm$0.14 & 0.0923$\pm$0.00 & 0.0512$\pm$0.05 & \textbf{0.2590$\pm$0.16} \\
CS & -0.0023$\pm$0.01 & 0.0088$\pm$0.04 & 0.0295$\pm$0.04 & 0.0250$\pm$0.03 & 0.0248$\pm$0.04 & \textbf{0.0647$\pm$0.03} & 0.0234$\pm$0.03 & 0.0040$\pm$0.02 & 0.0187$\pm$0.03 & -0.0142$\pm$0.00 & 0.0342$\pm$0.02 & \underline{0.0627$\pm$0.03} \\
HR & -0.0081$\pm$0.01 & -0.0008$\pm$0.01 & -0.0013$\pm$0.01 & -0.0010$\pm$0.01 & -0.0046$\pm$0.01 & -0.0033$\pm$0.01 & 0.0038$\pm$0.02 & -0.0064$\pm$0.01 & -0.0010$\pm$0.04 & 0.0094$\pm$0.00 & -0.0149$\pm$0.00 & \textbf{0.0368$\pm$0.01} \\
ZO & 0.5787$\pm$0.17 & 0.6085$\pm$0.17 & 0.5996$\pm$0.17 & 0.6091$\pm$0.15 & 0.6063$\pm$0.16 & 0.6094$\pm$0.12 & 0.6138$\pm$0.15 & 0.6143$\pm$0.15 & 0.4760$\pm$0.16 & \underline{0.7515$\pm$0.00} & 0.6442$\pm$0.07 & \textbf{0.7536$\pm$0.11} \\
BC & 0.0046$\pm$0.02 & 0.0240$\pm$0.05 & \textbf{0.0811$\pm$0.07} & 0.0777$\pm$0.07 & 0.0128$\pm$0.04 & 0.0523$\pm$0.05 & 0.0605$\pm$0.08 & 0.0692$\pm$0.08 & 0.0084$\pm$0.05 & 0.0393$\pm$0.00 & -0.0033$\pm$0.00 & \underline{0.0799$\pm$0.03} \\
LG & 0.0843$\pm$0.05 & 0.0951$\pm$0.03 & 0.1027$\pm$0.06 & 0.1432$\pm$0.07 & 0.1018$\pm$0.04 & 0.1544$\pm$0.09 & 0.1095$\pm$0.06 & 0.1262$\pm$0.05 & -0.0012$\pm$0.02 & 0.1293$\pm$0.00 & \textbf{0.1715$\pm$0.04} & \underline{0.1552$\pm$0.03} \\
TT & 0.0170$\pm$0.03 & 0.0184$\pm$0.02 & 0.0181$\pm$0.02 & 0.0095$\pm$0.01 & 0.0091$\pm$0.01 & 0.0156$\pm$0.01 & 0.0120$\pm$0.02 & 0.0124$\pm$0.02 & 0.0051$\pm$0.03 & -0.0022$\pm$0.00 & \underline{0.0198$\pm$0.02} & \textbf{0.0226$\pm$0.02} \\
AC & 0.2596$\pm$0.18 & \underline{0.3336$\pm$0.12} & 0.1210$\pm$0.15 & 0.1199$\pm$0.13 & 0.3092$\pm$0.11 & 0.3184$\pm$0.22 & 0.1913$\pm$0.18 & 0.1349$\pm$0.18 & 0.3263$\pm$0.19 & 0.0011$\pm$0.00 & 0.3107$\pm$0.11 & \textbf{0.4313$\pm$0.16} \\
VT & 0.5247$\pm$0.02 & 0.5463$\pm$0.01 & 0.5463$\pm$0.01 & \underline{0.5648$\pm$0.01} & 0.5368$\pm$0.01 & 0.5200$\pm$0.18 & 0.5572$\pm$0.00 & 0.5503$\pm$0.00 & 0.5367$\pm$0.00 & 0.0013$\pm$0.00 & 0.5572$\pm$0.00 & \textbf{0.6207$\pm$0.00} \\

CT4 & -0.0027$\pm$0.00 & 0.0010$\pm$0.00 & 0.0001$\pm$0.00 & 0.0004$\pm$0.00 & \underline{0.0013$\pm$0.00} & 0.0012$\pm$0.00 & 0.0006$\pm$0.00 & -0.0000$\pm$0.00 & -0.0005$\pm$0.00 & 0.0002$\pm$0.00 & 0.0005$\pm$0.00 & \textbf{0.0150$\pm$0.01} \\
OB & 0.1527$\pm$0.03 & 0.1588$\pm$0.02 & 0.1347$\pm$0.04 & 0.1522$\pm$0.04 & 0.1580$\pm$0.02 & 0.1548$\pm$0.04 & \textbf{0.1727$\pm$0.02} & 0.1417$\pm$0.05 & 0.1472$\pm$0.04 & 0.1527$\pm$0.04 & 0.1672$\pm$0.00 & \underline{0.1696$\pm$0.01} \\
AV & 0.0073$\pm$0.02 & 0.0111$\pm$0.02 & 0.0136$\pm$0.03 & \underline{0.0365$\pm$0.02} & 0.0123$\pm$0.03 & 0.0262$\pm$0.03 & 0.0214$\pm$0.03 & 0.0199$\pm$0.02 & 0.0173$\pm$0.02 & 0.0209$\pm$0.02 & 0.0166$\pm$0.00 & \textbf{0.0393$\pm$0.02} \\
BM & 0.0146$\pm$0.03 & 0.0107$\pm$0.02 & 0.0183$\pm$0.04 & 0.0054$\pm$0.05 & 0.0074$\pm$0.02 & 0.0178$\pm$0.02 & \underline{0.0186$\pm$0.01} & -0.0037$\pm$0.02 & 0.0035$\pm$0.02 & -0.0007$\pm$0.02 & 0.0072$\pm$0.00 & \textbf{0.0233$\pm$0.00} \\
CC & 0.0165$\pm$0.01 & 0.0166$\pm$0.01 & \underline{0.0189$\pm$0.02} & 0.0103$\pm$0.01 & 0.0150$\pm$0.01 & 0.0099$\pm$0.02 & 0.0019$\pm$0.01 & 0.0108$\pm$0.01 & 0.0113$\pm$0.02 & 0.0177$\pm$0.01 & 0.0106$\pm$0.00 & \textbf{0.0285$\pm$0.02} \\
CR & 0.0021$\pm$0.00 & 0.0033$\pm$0.00 & 0.0024$\pm$0.00 & 0.0039$\pm$0.00 & 0.0030$\pm$0.00 & \underline{0.0120$\pm$0.01} & 0.0114$\pm$0.00 & 0.0101$\pm$0.01 & 0.0106$\pm$0.01 & 0.0096$\pm$0.00 & 0.0122$\pm$0.00 & \textbf{0.0145$\pm$0.01} \\

\rone{HF} & \rone{0.0009$\pm$0.01} & \rone{-0.0025$\pm$0.01} & \rone{-0.0018$\pm$0.01} & \rone{-0.0015$\pm$0.01} & \rone{0.0024$\pm$0.02} & \rone{0.004$\pm$0.01} & \rone{-0.0029$\pm$0.01} & \rone{0.0005$\pm$0.01} & \rone{-0.0021$\pm$0.03} & \rone{\underline{0.0063$\pm$0.00} }& \rone{$-$} & \rone{\textbf{0.0072$\pm$0.01}} \\
        
\rone{MA} & \rone{0.1773$\pm$0.11} & \rone{\underline{0.2143$\pm$0.14}} & \rone{0.1140$\pm$0.13} & \rone{0.2048$\pm$0.13} & \rone{0.2047$\pm$0.14} & \rone{0.1586$\pm$0.07} & \rone{0.1983$\pm$0.08} & \rone{0.2035$\pm$0.08} & \rone{0.1259$\pm$0.12} & \rone{0.0120$\pm$0.00} & \rone{$-$} & \rone{\textbf{0.2165$\pm$0.06}} \\

\midrule
$\overline{\textit{AR}}$ & 8.50 & 6.40 & 6.85 & 6.28 & 6.78 & 5.40 & 5.85 & 7.40 & 8.72 & 8.20 & 6.38 & 1.25 \\
\bottomrule
\end{tabular}
}
\end{table*}

\textbf{20 benchmark datasets} in different fields from the UCI machine learning repository~\textcolor{blue}{\cite{uci}} have been selected to ensure a comprehensive evaluation. Their detailed statistical information is summarized in Table~\textcolor{blue}{\ref{tb:detaile_infor}}. All datasets are pre-processed by removing samples with missing values. In data ablation studies, we adopted a conservative nominal and ordinal attributes distinguishing protocol: attributes are defaulted to nominal unless exhibiting explicit semantic order (e.g., attribute ``Satisfaction'' with values $\{\text{very\_satisfied}, \text{satisfied}, $ $\text{average}, ...\}$). This mitigates subjective bias and crucially prevents distance structure distortion, which is particularly vital for methods that exploit semantic orders, such as DLC, ADC, and HARR. \rtwo{To ensure a fair comparison with existing methods, in all experiments, $k$ is set to the true number of clusters $k^*$ given by the dataset, because automatic determination of the number of clusters remains an open and challenging problem, especially for categorical and mixed data where basic distance metric is still under exploration.} 14 attributes of the SB dataset are omitted since each of them has only one possible value, which is meaningless for clustering. \rone{For brevity, possible values for the CT4 and CC datasets, which have too many attributes, are not listed in Table~\textcolor{blue}{\ref{tb:detaile_infor}}.} Since we focus on categorical data clustering, the numerical attributes of the datasets AP, DS, and AC are omitted, except for the performance evaluation on mixed data in Section~\textcolor{blue}{\ref{AP:CPMD}}. \rone{The MA dataset is derived by treating the integer-valued attributes in the AC dataset as possible values, and it is used together with the HF dataset to evaluate the OCL performance for attributes with numerous possible values.}

\begin{table*}[!t]
\caption{Clustering performance evaluated by NMI (with range [0,1]). ``$\overline{\textit{AR}}$'' row reports the average performance rankings.}
\label{tb:clustering_nm}
\centering
\resizebox{2.06\columnwidth}{!}{
\begin{tabular}{l|cccccccccccccc}
\toprule
Data & KMD & LSM & JDM & CBDM & UDMC & DLC & HDC & ADC & MCDC & AMPHM & HARR & OCL\\
\midrule
SB  & 0.8460$\pm$0.12 & 0.8418$\pm$0.13 & 0.8268$\pm$0.13 & 0.8393$\pm$0.12 & 0.8456$\pm$0.12 & 0.8298$\pm$0.12 & 0.8513$\pm$0.11 & 0.8317$\pm$0.10 & 0.5344$\pm$0.17 & 0.7892$\pm$0.00 & \underline{0.9157$\pm$0.14} & \textbf{0.9757$\pm$0.08} \\
NS  & 0.0458$\pm$0.03 & 0.0441$\pm$0.02 & 0.0412$\pm$0.02 & $-$ & 0.0492$\pm$0.03 & \underline{0.1348$\pm$0.10} & 0.0458$\pm$0.03 & 0.0458$\pm$0.03 & 0.0504$\pm$0.03 & $-$ & 0.0531$\pm$0.05 & \textbf{0.1492$\pm$0.10} \\
AP  & 0.0147$\pm$0.01 & 0.0321$\pm$0.03 & \underline{0.0746$\pm$0.03} & 0.0469$\pm$0.03 & 0.0276$\pm$0.02 & 0.0649$\pm$0.00 & 0.0384$\pm$0.04 & 0.0131$\pm$0.01 & 0.0194$\pm$0.02 & 0.0004$\pm$0.00 & 0.0185$\pm$0.03 & \textbf{0.0955$\pm$0.01} \\
DS  & 0.1352$\pm$0.14 & 0.1352$\pm$0.14 & 0.2552$\pm$0.21 & 0.2552$\pm$0.21 & 0.2034$\pm$0.20 & 0.1682$\pm$0.21 & 0.2034$\pm$0.20 & 0.2082$\pm$0.20 & \textbf{0.3083$\pm$0.13} & 0.0646$\pm$0.00 & 0.0623$\pm$0.04 & \underline{0.2732$\pm$0.18} \\
CS  & 0.0157$\pm$0.03 & 0.0221$\pm$0.04 & 0.0426$\pm$0.05 & 0.0353$\pm$0.03 & 0.0413$\pm$0.05 & \textbf{0.0840$\pm$0.03} & 0.0346$\pm$0.03 & 0.0155$\pm$0.02 & 0.0310$\pm$0.03 & 0.0034$\pm$0.00 & 0.0344$\pm$0.01 & \underline{0.0808$\pm$0.04} \\
HR  & 0.0088$\pm$0.01 & 0.0144$\pm$0.01 & 0.0133$\pm$0.01 & 0.0110$\pm$0.01 & 0.0091$\pm$0.01 & 0.0100$\pm$0.01 & 0.0182$\pm$0.02 & 0.0080$\pm$0.01 & \underline{0.0407$\pm$0.08} & 0.0304$\pm$0.00 & 0.0000$\pm$0.00 & \textbf{0.0456$\pm$0.02} \\
ZO  & 0.7172$\pm$0.08 & 0.7438$\pm$0.09 & 0.7395$\pm$0.09 & 0.7439$\pm$0.08 & 0.7541$\pm$0.09 & 0.7799$\pm$0.06 & 0.7620$\pm$0.07 & 0.7731$\pm$0.08 & 0.7057$\pm$0.10 & 0.7190$\pm$0.00 & \underline{0.8324$\pm$0.04} & \textbf{0.8332$\pm$0.04} \\
BC  & 0.0044$\pm$0.01 & 0.0123$\pm$0.02 & \underline{0.0355$\pm$0.03} & \textbf{0.0356$\pm$0.03} & 0.0068$\pm$0.01 & 0.0195$\pm$0.01 & 0.0282$\pm$0.04 & 0.0318$\pm$0.03 & 0.0156$\pm$0.02 & 0.0091$\pm$0.00 & 0.0017$\pm$0.00 & 0.0281$\pm$0.01 \\
LG  & 0.1309$\pm$0.04 & 0.1490$\pm$0.04 & 0.1477$\pm$0.05 & 0.1885$\pm$0.07 & 0.1537$\pm$0.04 & 0.1921$\pm$0.07 & 0.1620$\pm$0.05 & 0.1680$\pm$0.03 & 0.0832$\pm$0.06 & \textbf{0.3300$\pm$0.00} & \underline{0.2047$\pm$0.04} & 0.1919$\pm$0.04 \\
TT  & 0.0105$\pm$0.01 & \underline{0.0148$\pm$0.01} & 0.0146$\pm$0.02 & 0.0084$\pm$0.01 & 0.0074$\pm$0.01 & 0.0078$\pm$0.00 & 0.0092$\pm$0.01 & 0.0094$\pm$0.01 & 0.0075$\pm$0.01 & 0.0008$\pm$0.00 & \textbf{0.0163$\pm$0.02} & 0.0132$\pm$0.02 \\
AC  & 0.2208$\pm$0.13 & 0.2686$\pm$0.10 & 0.0935$\pm$0.11 & 0.0920$\pm$0.10 & 0.2430$\pm$0.08 & 0.2677$\pm$0.19 & 0.1667$\pm$0.11 & 0.1266$\pm$0.12 & \underline{0.2803$\pm$0.12} & 0.0009$\pm$0.00 & 0.2402$\pm$0.08 & \textbf{0.3643$\pm$0.14} \\
VT  & 0.4575$\pm$0.03 & 0.4808$\pm$0.02 & 0.4808$\pm$0.02 & \underline{0.4948$\pm$0.00} & 0.4673$\pm$0.02 & 0.4541$\pm$0.16 & 0.4893$\pm$0.00 & 0.4844$\pm$0.00 & 0.4747$\pm$0.00 & 0.0001$\pm$0.00 & 0.4893$\pm$0.00 & \textbf{0.5322$\pm$0.00} \\

CT4 & 0.0028$\pm$0.00 & 0.0029$\pm$0.00 & 0.0017$\pm$0.00 & 0.0015$\pm$0.00 & \underline{0.0036$\pm$0.00} & 0.0013$\pm$0.00 & 0.0011$\pm$0.00 & 0.0016$\pm$0.00 & 0.0016$\pm$0.00 & 0.0053$\pm$0.00 & 0.0023$\pm$0.00 & \textbf{0.0128$\pm$0.01} \\
OB & 0.2256$\pm$0.03 & 0.2330$\pm$0.02 & 0.2081$\pm$0.04 & 0.2262$\pm$0.03 & 0.2342$\pm$0.02 & 0.2391$\pm$0.03 & \textbf{0.2644$\pm$0.02} & 0.2261$\pm$0.04 & 0.2233$\pm$0.03 & 0.2466$\pm$0.03 & 0.2333$\pm$0.00 & \underline{0.2633$\pm$0.01} \\
AV & 0.0021$\pm$0.00 & 0.0024$\pm$0.00 & 0.0024$\pm$0.00 & \underline{0.0099$\pm$0.01} & 0.0027$\pm$0.00 & 0.0093$\pm$0.01 & 0.0075$\pm$0.01 & 0.0035$\pm$0.00 & 0.0034$\pm$0.00 & 0.0025$\pm$0.01 & 0.0044$\pm$0.00 & \textbf{0.0110$\pm$0.01} \\
BM & \textbf{0.0211$\pm$0.00} & 0.0143$\pm$0.01 & 0.0135$\pm$0.01 & 0.0198$\pm$0.01 & 0.0179$\pm$0.01 & 0.0121$\pm$0.01 & 0.0155$\pm$0.01 & 0.0192$\pm$0.01 & \underline{0.0204$\pm$0.01} & 0.0103$\pm$0.01 & 0.0177$\pm$0.00 & 0.0194$\pm$0.00 \\
CC & 0.0150$\pm$0.01 & 0.0149$\pm$0.01 & 0.0149$\pm$0.01 & 0.0084$\pm$0.01 & 0.0119$\pm$0.01 & 0.0076$\pm$0.01 & 0.0023$\pm$0.01 & 0.0148$\pm$0.01 & 0.0141$\pm$0.02 & \underline{0.0156$\pm$0.01} & 0.0144$\pm$0.00 & \textbf{0.0213$\pm$0.02} \\
CR & 0.0017$\pm$0.00 & 0.0024$\pm$0.00 & 0.0020$\pm$0.00 & 0.0022$\pm$0.00 & 0.0022$\pm$0.00 & 0.0071$\pm$0.01 & \textbf{0.0115$\pm$0.00} & 0.0065$\pm$0.00 & 0.0057$\pm$0.00 & 0.0066$\pm$0.00 & 0.0048$\pm$0.00 & \underline{0.0083$\pm$0.01} \\
     
\rone{HF}& \rone{0.0023$\pm$0.03} & \rone{0.0016$\pm$0.03} & \rone{0.0013$\pm$0.05} & \rone{0.0016$\pm$0.04} & \rone{0.0023$\pm$0.03} & \rone{\textbf{0.0041$\pm$0.03}} & \rone{0.0021$\pm$0.03} & \rone{\underline{0.0026$\pm$0.01}} & \rone{0.0018$\pm$0.03} & \rone{0.0019$\pm$0.00} & \rone{$-$} & \rone{\underline{0.0026$\pm$0.01}} \\
        
\rone{MA}& \rone{0.1575$\pm$0.08} & \rone{\textbf{0.1705$\pm$0.10}} & \rone{0.0939$\pm$0.09} & \rone{\underline{0.1700$\pm$0.10}} & \rone{\underline{0.1700$\pm$0.10}} & \rone{0.1692$\pm$0.15} & \rone{0.1464$\pm$0.05} & \rone{0.1523$\pm$0.05} & \rone{0.1079$\pm$0.08} & \rone{0.0173$\pm$0.00} & \rone{$-$} & \rone{0.1582$\pm$0.03} \\

\midrule
$\overline{\textit{AR}}$ &  \rone{7.84} &  \rone{6.74} &  \rone{7.61} &  \rone{6.55} &  \rone{6.53} &  \rone{5.47} &  \rone{6.03} &  \rone{6.84} &  \rone{7.24} &  \rone{8.13} &  \rone{6.84} &  \rone{2.18} \\
\bottomrule
\end{tabular}}
\end{table*}

\rtwo{\textbf{11 counterparts} sorted out in Table~\textcolor{blue}{\ref{tb:method_statistics}} have been selected for the comparative study.} The counterparts include traditional K-MoDes (KMD) algorithm~\textcolor{blue}{\cite{kmd}}, popular entropy-based metric LSM~\textcolor{blue}{\cite{lsm}}, and context-based distance measurement CBDM~\textcolor{blue}{\cite{cbdm_conf}},  statistical-based Jia's Distance Metric (JDM)~\textcolor{blue}{\cite{jdm}} and Unified Distance Metric based Clustering (UDMC)~\textcolor{blue}{\cite{udm}}, state-of-the-art HD based Clustering (HDC)~\textcolor{blue}{\cite{HDC}}, Heterogeneous Attribute Reconstruction and Representation based clustering (HARR)~\textcolor{blue}{\cite{HARR}}, Distance Learning based Clustering (DLC) algorithm~\textcolor{blue}{\cite{dlc}}, graph metric based clustering algorithm ADC~\textcolor{blue}{\cite{ADC}}, Multi-granular-guided Categorical Data Clustering (MCDC) algorithm~\textcolor{blue}{\cite{ICDCSCai}}, and Adaptive Micro Partition and Hierarchical Merging (AMPHM) clustering algorithm~\textcolor{blue}{\cite{CAIS}}. Among them, LSM and CBDM are traditional distance measures for categorical data, while JDM and UDMC are statistical-based measures. All four are adopted by KMD to perform clustering. DLC, HDC, ADC, MCDC, AMPHM, and HARR are state-of-the-art categorical data clustering methods. 
The parameters (if any) of these methods are set to the values recommended in the corresponding paper. Each clustering result was recorded as the average performance with standard deviations of 10 runs of the compared methods.

\begin{table*}[!t]
\caption{\rtwo{Clustering performance evaluated by CMP (with range $[0,1]$). ``$\overline{\textit{AR}}$'' row reports the average performance rankings.}}
\label{tb:clustering_cmp}
\centering
\resizebox{2.06\columnwidth}{!}{
\begin{tabular}{c|cccccccccccc}
\toprule
Datasets & KMD & LSM & JDM & CBDM & UDMC & DLC & HDC & ADC & MCDC & AMPHM & HARR & OCL \\
\midrule
SB & 0.4021$\pm$0.05 & 0.3966$\pm$0.04 & 0.3931$\pm$0.04 & 0.3909$\pm$0.04 & 0.3873$\pm$0.03 & 0.3960$\pm$0.03 & 0.3829$\pm$0.03 & \underline{0.3824$\pm$0.04} & 0.4801$\pm$0.07 & 0.3832$\pm$0.00 & 0.4743$\pm$0.02 & \textbf{0.3659$\pm$0.01} \\
NS & 0.8917$\pm$0.01 & 0.8768$\pm$0.01 & 0.8405$\pm$0.01 & $-$ & 0.8799$\pm$0.01 & \underline{0.7855$\pm$0.03} & 0.8917$\pm$0.01 & 0.8917$\pm$0.01 & 0.8036$\pm$0.03 & $-$ & 0.8830$\pm$0.06 & \textbf{0.7796$\pm$0.01} \\
AP & 0.6067$\pm$0.01 & 0.6021$\pm$0.02 & 0.5868$\pm$0.02 & 0.5989$\pm$0.02 & 0.6060$\pm$0.01 & 0.5743$\pm$0.00 & 0.5950$\pm$0.02 & \textbf{0.5731$\pm$0.09} & 0.5882$\pm$0.05 & 0.6211$\pm$0.00 & 0.6885$\pm$0.04 & \underline{0.5737$\pm$0.00} \\
DS & 0.6535$\pm$0.05 & 0.6535$\pm$0.05 & 0.6186$\pm$0.05 & 0.6136$\pm$0.05 & \textbf{0.6068$\pm$0.04} & 0.6590$\pm$0.06 & \underline{0.6068$\pm$0.04} & 0.6177$\pm$0.05 & 0.6363$\pm$0.13 & 0.7084$\pm$0.00 & 0.6906$\pm$0.09 & 0.6174$\pm$0.06 \\
CS & 0.7428$\pm$0.02 & 0.7374$\pm$0.05 & 0.7152$\pm$0.06 & 0.7243$\pm$0.04 & 0.7242$\pm$0.04 & \underline{0.6620$\pm$0.03} & 0.7477$\pm$0.01 & 0.7506$\pm$0.01 & 0.6982$\pm$0.09 & 0.7272$\pm$0.00 & 0.7060$\pm$0.05 & \textbf{0.6618$\pm$0.03} \\
HR & 0.7138$\pm$0.01 & 0.7111$\pm$0.01 & 0.7136$\pm$0.02 & 0.7162$\pm$0.01 & 0.7106$\pm$0.01 & \underline{0.6797$\pm$0.01} & 0.7196$\pm$0.02 & 0.7128$\pm$0.01 & 0.7682$\pm$0.09 & 0.7009$\pm$0.00 & \textbf{0.6297$\pm$0.00} & 0.6994$\pm$0.01 \\
ZO & 0.2823$\pm$0.02 & 0.2854$\pm$0.03 & 0.2809$\pm$0.02 & \underline{0.2801$\pm$0.02} & 0.2864$\pm$0.03 & 0.2832$\pm$0.02 & 0.2864$\pm$0.03 & 0.2886$\pm$0.03 & 0.3454$\pm$0.05 & 0.2944$\pm$0.00 & 0.2812$\pm$0.01 & \textbf{0.2799$\pm$0.02} \\
BC & 0.7163$\pm$0.01 & 0.7180$\pm$0.01 & 0.7139$\pm$0.02 & 0.7174$\pm$0.02 & 0.7172$\pm$0.01 & 0.6960$\pm$0.01 & 0.7201$\pm$0.01 & 0.7137$\pm$0.02 & \underline{0.6858$\pm$0.02} & 0.7554$\pm$0.00 & 0.7547$\pm$0.01 & \textbf{0.6801$\pm$0.01} \\
LG & 0.6155$\pm$0.01 & 0.5945$\pm$0.01 & 0.5996$\pm$0.02 & 0.5985$\pm$0.01 & 0.6017$\pm$0.02 & 0.5797$\pm$0.01 & 0.5826$\pm$0.02 & 0.5804$\pm$0.03 & \textbf{0.5470$\pm$0.03} & \underline{0.5649$\pm$0.00} & 0.6790$\pm$0.01 & 0.5781$\pm$0.01 \\
TT & 0.9213$\pm$0.00 & 0.9217$\pm$0.01 & 0.9220$\pm$0.01 & 0.9214$\pm$0.00 & 0.9216$\pm$0.00 & \underline{0.8905$\pm$0.00} & 0.9223$\pm$0.00 & 0.9222$\pm$0.00 & 0.8922$\pm$0.00 & 0.9275$\pm$0.00 & 0.9172$\pm$0.00 & \textbf{0.8876$\pm$0.00} \\
AC & 0.6560$\pm$0.04 & \underline{0.6448$\pm$0.03} & 0.6605$\pm$0.03 & 0.6644$\pm$0.04 & 0.6602$\pm$0.02 & 0.6462$\pm$0.02 & 0.6707$\pm$0.03 & 0.6800$\pm$0.02 & 0.6468$\pm$0.01 & 0.7878$\pm$0.00 & 0.8044$\pm$0.01 & \textbf{0.6356$\pm$0.03} \\
VT & 0.5510$\pm$0.00 & 0.5511$\pm$0.00 & 0.5511$\pm$0.00 & 0.5464$\pm$0.00 & 0.5482$\pm$0.00 & 0.5633$\pm$0.06 & \underline{0.5450$\pm$0.00} & 0.5451$\pm$0.00 & 0.5458$\pm$0.00 & 0.5772$\pm$0.00 & 0.5454$\pm$0.00 & \textbf{0.5427$\pm$0.00} \\
CT4 & 0.401$\pm$0.01&	0.4004$\pm$0.01&	0.3914$\pm$0.01&	0.3906$\pm$0.01&	0.3978$\pm$0.01	& \textbf{0.3811$\pm$0.01}	& \underline{0.3858$\pm$0.01}&0.3872$\pm$0.01	&0.3988$\pm$0.02 & 0.4013$\pm$ 0.00 &0.3968$\pm$0.01 &0.3952$\pm$0.01 \\
OB & 0.3403$\pm$0.05	&0.3359$\pm$0.04	&0.3887$\pm$0.05	&0.3752$\pm$0.06	&0.3639$\pm$0.06	&0.3280$\pm$0.03	&0.3825$\pm$0.06	&0.3688$\pm$0.04 &0.3663$\pm$0.03 & 0.3841$\pm$0.00 & \underline{0.3019$\pm$0.02}    & \textbf{0.2973$\pm$0.01} \\
AV & 0.7039$\pm$0.02&	0.6861$\pm$0.02&	\textbf{0.6265$\pm$0.04}&	0.6623$\pm$0.05&	0.6912$\pm$0.02&	\underline{0.6355$\pm$0.02}& 0.7011$\pm$0.03	&0.6844$\pm$0.05	&0.6760$\pm$0.02 & 0.6585$\pm$0.00 & 0.6987 $\pm$ 0.07&0.6518$\pm$0.02 \\
BM & 0.6153$\pm$0.01&	0.6205$\pm$0.03&	0.6255$\pm$0.02&	0.6144$\pm$0.02&	0.6152$\pm$0.02&	\underline{0.6017$\pm$0.03} &	0.6281$\pm$0.01&	0.6242$\pm$0.01	&0.6128$\pm$0.03 &0.7326$\pm$0.00 &0.7875$\pm$0.01 & \textbf{0.5762$\pm$0.00} \\
CC &0.5342$\pm$0.01&	0.5343$\pm$0.01	&0.5355$\pm$0.01	&0.5334$\pm$0.02	&0.5333$\pm$0.01&	0.5466$\pm$0.01&	\underline{0.5322$\pm$0.03}	&0.5354$\pm$0.02	&0.5426$\pm$0.03 &0.5595$\pm$0.00 & 0.5637$\pm$0.03 & \textbf{0.5303$\pm$0.01} \\
CR &0.8339$\pm$0.01	&0.8278$\pm$0.01	&0.8238$\pm$0.02	&0.8288$\pm$0.02	&0.8257$\pm$0.01	&0.7880$\pm$0.04 & 0.8047$\pm$0.01&	0.8008$\pm$0.01	& \underline{0.7826$\pm$0.03} &0.8045$\pm$0.00 & 0.8645 $\pm$ 0.00& \textbf{0.7795$\pm$0.02}\\
HF & 0.8257$\pm$0.01&	0.8174$\pm$0.01&	0.8118$\pm$0.01&	0.8140$\pm$0.01&	0.8184$\pm$0.01&	0.8187$\pm$0.01&	0.8297$\pm$0.01&	0.8518$\pm$0.01& \underline{0.7960$\pm$0.07} &0.8244$\pm$0.00 &$-$ & \textbf{0.7942$\pm$0.01} \\
MA & 0.6857$\pm$0.02	&0.6865$\pm$0.02&	0.6925$\pm$0.02&	0.6850$\pm$0.02&	0.6850$\pm$0.01&	0.6957$\pm$0.01&	\textbf{0.6429$\pm$0.01}&	0.6815$\pm$0.01&0.6861$\pm$0.01 &0.7289$\pm$0.00& $-$ & \underline{0.6690$\pm$0.01}
 \\

\midrule
$\overline{\textit{AR}}$ &  \rtwo{8.07} & \rtwo{7.25} & \rtwo{6.53} & \rtwo{6.35} & \rtwo{6.38} & \rtwo{4.70} & \rtwo{7.00} & \rtwo{6.45} & \rtwo{5.70} & \rtwo{9.18} & \rtwo{8.60} & \rtwo{1.80} \\
\bottomrule
\end{tabular}
}
\end{table*}

\rtwo{\textbf{Three external and one internal validity indices} are employed to evaluate clustering performance. The external indices: Clustering Accuracy (CA) \textcolor{blue}{\cite{ex1}}, Adjusted Rand Index (ARI) \textcolor{blue}{\cite{ex4,ex3}} and Normalized Mutual Information (NMI) \textcolor{blue}{\cite{ex5}} measure alignment with ground truth, where higher values indicate better performance. The internal index, an entropy-based Clustering coMPactness (CMP) is introduced to eliminate the influence of different distance metric magnitudes, which is computed as:}
\begin{equation}
\label{eq:cmp}
\text{CMP} = \frac{1}{s \times k} \sum_{j=1}^{k} \sum_{r=1}^{s} \sum_{g=1}^{V_r}\frac{ -p(v_{r,g} | C_j)\log p(v_{r,g} | C_j)}{\log l_r}.
\end{equation}
\rtwo{Lower CMP indicates higher value concentration of samples within clusters, revealing better cluster compactness. In addition, the Wilcoxon signed rank test~\textcolor{blue}{\cite{wilcoxon}} is also performed to validate the performance difference between OCL and the competitive counterparts.}

\subsection{Clustering Performance Evaluation}
\label{sct:CPE}

\rone{The clustering performance of various methods is evaluated as shown in Tables~\textcolor{blue}{\ref{tb:clustering_ca}} - \textcolor{blue}{\ref{tb:clustering_cmp}}. The best and second-best results on each dataset are highlighted in \textbf{bold} and \underline{underlined}, respectively. }
The results of CBDM and AMPHM are not reported on the NS dataset because all the NS's attributes are independent of each other, making the inter-attribute dependence-based methods like CBDM and AMPHM fail in measuring distances. \rone{Moreover, the clustering performance of HARR on the datasets HF and MA is not reported because the projection mechanism of HARR generates excessive endogenous subspaces when handling multiple categorical values, resulting in computationally intractable overhead.}

\begin{table}[!t]
\centering
\caption{\rtwo{Wilcoxon signed-rank test comparing OCL vs. the best non-OCL method on each dataset based on CA performance. The \textbf{Results} column indicates whether the difference is significant at the 95\% confidence level.}}
 
\resizebox{0.99\columnwidth}{!}{
\begin{tabular}{l|lcc||l|lcc}
\toprule
 \textbf{Data} &\textbf{Counterpart} & \textbf{p-value} & \textbf{Results} & \textbf{Data} &\textbf{Counterpart} & \textbf{p-value} & \textbf{Results}\\
\midrule
SB & vs. HARR  &  0.31731& $\times$ & AC &vs. LSM  &  0.00195& \checkmark \\

 NS & vs. DLC  & 0.01367 & \checkmark & VT & vs. CBDM  &  0.00195 & \checkmark\\
 
 AP & vs. JDM  & 0.75183 & $\times$& CT4 & vs. KMD  &  0.19335 & $\times$  \\
 
DS & vs. CBDM  &  0.02601 & \checkmark & OB & vs. HDC  & 0.00390 & \checkmark \\

CS & vs. DLC&  0.17971 & $\times$ & AV & vs. CBDM & 0.17320 & $\times$  \\

HR &  vs. AMPHM  & 0.00195& \checkmark & BM & vs. HDC  &  0.00977 & \checkmark \\

ZO & vs. HARR  &  0.00390& \checkmark & CC & vs. AMPHM  &  0.02734 & \checkmark\\

BC &vs. AMPHM  & 0.00977 & \checkmark & CR & vs. HDC  & 0.00195 & \checkmark\\

LG & vs. AMPHM  & 0.00195 & \checkmark &  HF &  vs. AMPHM  & 0.00195 & \checkmark \\
TT &  vs. MCDC & 0.05041& $\times$ & MA &  vs. UDMC  &  0.03389 & \checkmark\\

\bottomrule
\end{tabular}} \label{tb:wtcc}
\end{table}

\begin{table*}[!t]
\centering
\caption{\rtwo{Clustering performance of OCL and its ablation variants. OCL-I, OCL-II, and OCL-III are ablated variants of OCL, obtained by successively removing the consideration of order distance, order learning, and order information from OCL. For more detailed description of the ablation variants, please refer to the main text of Section~\textcolor{blue}{\ref{sct:ablation}}.}}
\label{tb:ab_ca}
\centering
\resizebox{2.06\columnwidth}{!}{
\begin{tabular}{c|cccc|cccc|cccc|cccc}
\toprule
\multirow{2}{*}{Datasets} & \multicolumn{4}{c}{CA} & \multicolumn{4}{c}{ARI}& \multicolumn{4}{c}{NMI}& \multicolumn{4}{c}{\rtwo{CMP}} \\
\cmidrule(lr){2-5}\cmidrule(lr){6-9}\cmidrule(lr){10-13}\cmidrule(lr){14-17}
 & OCL-III & OCL-II & OCL-I & OCL  & OCL-III & OCL-II & OCL-I & OCL  & OCL-III & OCL-II & OCL-I & OCL& \rtwo{OCL-III} & \rtwo{OCL-II} & \rtwo{OCL-I} & \rtwo{OCL}\\
\midrule
SB & 0.8468 & 0.8681 & 0.9149 & \textbf{0.9830}  & 0.7638 & 0.8371 & 0.8718 & \textbf{0.9620} & 0.8460 & 0.9113 & 0.9262 & \textbf{0.9757}& \rtwo{0.4021} 	& \rtwo{0.3854} & \rtwo{0.3770} & \rtwo{\textbf{0.3659}}\\
NS & 0.3208 & 0.3359 & 0.3364 & \textbf{0.3573} & 0.0376 & 0.0614 & 0.0592 & \textbf{0.1015}&0.0458 & 0.1122 & 0.1086 & \textbf{0.1492}& \rtwo{0.8917} 	& \rtwo{0.7915} 	& \rtwo{0.7915} &	\rtwo{\textbf{0.7796}} 
\\
AP & 0.5392 & 0.6069 & 0.6085 & \textbf{0.6222}& 0.0022 & 0.0384 &  0.0388 & \textbf{0.0524} &  0.0147 & 0.0744 &  0.0919 & \textbf{0.0955}& \rtwo{0.6067} &	\rtwo{0.6011} &	\rtwo{0.5937} &	\rtwo{\textbf{0.5737} }
\\
DS & 0.6525 & 0.7125 & 0.7283 & \textbf{0.7458} & 0.1077 & 0.2295& 0.2407 & \textbf{0.2590}&  0.1352 & 0.2462 & 0.2567 & \textbf{0.2732}& \rtwo{0.6535} & \rtwo{0.6401} &	\rtwo{0.6392}& 	\rtwo{\textbf{0.6174 }}
\\
CS & 0.5463 & 0.6388 & \textbf{0.6500} & 0.6313 &  -0.0023 & 0.0670 & \textbf{0.0780} & 0.0627&  0.0157 & 0.0816 & \textbf{0.0975} & 0.0808& \rtwo{0.7428} & \rtwo{0.6695} & \rtwo{\textbf{0.6572}} & \rtwo{0.6618}\\
HR & 0.3758 & 0.3712 & 0.3765 & \textbf{0.4326} &  -0.0081 & -0.0003 & 0.0079 & \textbf{0.0368}&  0.0088 & 0.0153 & 0.0236 & \textbf{0.0456}& \rtwo{0.7138} & \rtwo{0.6953} & \rtwo{0.7040} & \rtwo{\textbf{0.6994}}\\
ZO & 0.6624 & 0.7000 & 0.7525 & \textbf{0.7792}  & 0.5787 & 0.6482 & 0.7160 & \textbf{0.7536}&  0.7172 & 0.7965 & 0.8111 & \textbf{0.8332}& \rtwo{0.2823} & \rtwo{0.2968} & \rtwo{0.2906} & \rtwo{\textbf{0.2799}}\\
BC & 0.5472 & 0.5636 & 0.5643 & \textbf{0.6650}  &0.0046 & 0.0247 & 0.0248 & \textbf{0.0799}&  0.0044 & 0.0110 & 0.0112 & \textbf{0.0281}& \rtwo{0.7163} & \rtwo{0.6953} & \rtwo{0.6930} & \rtwo{\textbf{0.6801}}\\
LG & 0.4277 & 0.4453 & 0.4872 & \textbf{0.5426} & 0.0843 & 0.0953 & 0.1054 & \textbf{0.1552}& 0.1309 & 0.1410 & 0.1572 & \textbf{0.1919}& \rtwo{0.6155} & \rtwo{0.5983} & \rtwo{0.5892} & \rtwo{\textbf{0.5781}}\\
TT & 0.5569 & 0.5762 & 0.5784 & \textbf{0.5785} &  0.0170 & 0.0205 & 0.0132 & \textbf{0.0226}& 0.0105 & 0.0106 & 0.0062 & \textbf{0.0132}& \rtwo{0.9213} & \rtwo{0.8978} & \rtwo{0.8886} & \rtwo{\textbf{0.8876}}\\
AC & 0.7281 & 0.7545 & 0.7546 & \textbf{0.8206}& 0.2596 & 0.2879 & 0.2880 & \textbf{0.4313}&  0.2208 & 0.2237 &  0.2238 & \textbf{0.3643} & \rtwo{0.6560} & \rtwo{0.6418} & \rtwo{0.6418} & \rtwo{\textbf{0.6356}}\\
VT & 0.8625 & 0.8828 & 0.8942 & \textbf{0.8943} & 0.5247 & 0.5850 & \textbf{0.6208} &0.6207& 0.4575 &0.5098 & \textbf{0.5422} & 0.5322& \rtwo{0.5510} & \rtwo{0.5451} & \rtwo{0.5443} & \rtwo{\textbf{0.5427}}\\
CT4 & 0.4000 &0.3612 &  0.4034 & \textbf{0.4121} & -0.0027 & 0.0018 & 0.0024 & \textbf{0.0150} & 0.0028 & 0.0019 & 0.0078  & \textbf{0.0128}& \rtwo{0.4010} & \rtwo{0.3991} & \rtwo{0.3994} & \rtwo{\textbf{0.3952}}\\
OB & 0.3715 & 0.3596 & 0.3910 & \textbf{0.3935} & 0.1527 & 0.1410 & \textbf{0.1814} & 0.1696 & 0.2256 & 0.2433 & 0.2572 & \textbf{0.2633}& \rtwo{0.3403} & \rtwo{0.2992} & \rtwo{0.3036} & \rtwo{\textbf{0.2973}}\\
AV & 0.6251 & \textbf{0.7144} & \textbf{0.7144} & 0.6637  & 0.0073 & \textbf{0.0483} & \textbf{0.0483} & 0.0393& 0.0021 & 0.0088 & 0.0088 & \textbf{0.0110}& \rtwo{0.7039} & \rtwo{\textbf{0.6346 }}& \rtwo{\textbf{0.6346}} & \rtwo{0.6518}\\
BM & 0.5730 & 0.6029 & 0.6049 & \textbf{0.6127} & 0.0146  & 0.0158 & 0.0165 & \textbf{0.0233} & \textbf{0.0211} & 0.0166 & 0.0181  & 0.0194& \rtwo{0.6153} & \rtwo{0.5859} & \rtwo{0.5802} & \rtwo{\textbf{0.5762}}\\
CC & 0.5562 & 0.5710 & 0.5711 & \textbf{0.5726} & 0.0165 & 0.0284 & \textbf{0.0285 }& \textbf{0.0285} & 0.0150 & 0.0211 & 0.0212 & \textbf{0.0213}& \rtwo{0.5342} & \rtwo{0.5298} & \rtwo{\textbf{0.5295}} & \rtwo{0.5303}\\
CR & 0.5209 & 0.5507 & 0.5512 & \textbf{0.5590} & 0.0021 & 0.0121 & 0.0128 & \textbf{0.0145}& 0.0017 & 0.0072 & 0.0080 & \textbf{0.0083}& \rtwo{0.8339} & \rtwo{0.7807} & \rtwo{0.7852} & \rtwo{\textbf{0.7795}}\\
\rone{HF} & \rone{0.5482} & \rone{0.5331} & \rone{0.5495}  & \rone{\textbf{0.5652}}& \rone{0.0009}  & \rone{0.0002} & \rone{0.0021}& \rone{\textbf{0.0072}}  & \rone{0.0023} & \rone{0.0013} & \rone{0.0008} & \rone{\textbf{0.0026}}& \rone{0.8257} & \rone{0.8042} & \rone{0.7947} & \rone{\textbf{0.7942}}\\
\rone{MA} & \rone{0.6961} & \rone{\textbf{0.7430}} & \rone{\textbf{0.7430}} & \rone{0.7333} & \rone{0.1773} & \rone{\textbf{0.2622}} & \rone{\textbf{ 0.2622}} & \rone{0.2165} & \rone{0.1575} & \rone{\textbf{0.2019}} & \rone{\textbf{0.2019}} & \rone{0.1582} & \rone{0.6857} & \rone{0.6771} & \rone{0.6766} & \rone{\textbf{0.6690}}\\
\midrule
$\overline{\textit{AR}}$ & \rtwo{3.80} &  \rtwo{3.00} &  \rtwo{1.90} &  \rtwo{1.30} &  \rtwo{3.85} &  \rtwo{2.80} &  \rtwo{1.93} &  \rtwo{1.43}&  \rtwo{3.65} & \rtwo{2.85} & \rtwo{2.20} & \rtwo{1.30}   & \rtwo{3.90} & \rtwo{2.62} & \rtwo{2.17} & \rtwo{1.30} \\
\bottomrule
\end{tabular}}
\end{table*}

\rthree{\textbf{External (label-based) validation:} The clustering results in terms of different validity indices are shown in Tables \textcolor{blue}{\ref{tb:clustering_ca}} - \textcolor{blue}{\ref{tb:clustering_nm}}.} The observations include the following five aspects: (1) Overall, OCL reaches the best result on most datasets, indicating its superiority in clustering. (2) On the CA index, although OCL does not perform significantly better than the second-best method on the AP and CR datasets, the second-best methods are different on these datasets, but OCL has stable performance on them. (3) On the CA index, although OCL does not perform the best on CS, AV and TT datasets, its performance is close to the best result (the difference is less than 0.005 in CS and AV datasets, less than 0.03 in TT dataset), which still proves OCL's competitiveness. (4) In terms of the ARI index, there are five datasets on which OCL does not perform the best, namely AP, CS, BC, LG, and OB. However, the differences between OCL and the best-performing algorithms on AP, CS, and BC datasets are very small. (5) \rthree{As} for the NMI index, although OCL does not have the best NMI performance on some datasets, it is not surpassed by much by the winners and is still very competitive.

\rtwo{\textbf{Internal validation}: As can be seen in Table \textcolor{blue}{\ref{tb:clustering_cmp}}, the proposed OCL still performs the best in general in exploring compact and meaningful clusters, as it infers the ordinal structure of attribute values and finely adjusts the distance metric according to each cluster to better serve the clustering objective. 
}


\rtwo{\textbf{Significant studies:} Table~\textcolor{blue}{\ref{tb:wtcc}} demonstrates the Wilcoxon signed rank test~\textcolor{blue}{\cite{wilcoxon}} results based on the CA results reported in Table~\textcolor{blue}{\ref{tb:clustering_ca}}. For each dataset, OCL is compared with the best-performing counterpart using the one-sided Wilcoxon test under a 95\% confidence level. It can be observed that OCL significantly outperforms the second-best method in most cases, providing strong statistical evidence for its overall superiority. Moreover, on some datasets where OCL does not perform the best (i.e., CS, TT, and AV), its performance is not significantly worse than the best-performing counterparts, demonstrating the competitiveness of OCL.}

\begin{table*}[!t]
\centering
\caption{\rtwo{Clustering performance of OCL variants with differential treatment of nominal and ordinal attributes: LNRO learns orders only for nominal attributes, while RNRO preserves the semantic order of ordinal attributes and does not perform order learning for all attributes. For more detailed description of the two OCL variants, please refer to the main text of Section~\textcolor{blue}{\ref{sct:ablation}}.}}
\label{tb:ab_ca_d}
\centering
\resizebox{1.8\columnwidth}{!}{
\begin{tabular}{c|ccc|ccc|ccc|ccc}
\toprule
\multirow{2}{*}{Datasets} & \multicolumn{3}{c}{CA} & \multicolumn{3}{c}{ARI}& \multicolumn{3}{c}{NMI}& \multicolumn{3}{c}{\rtwo{CMP}} \\
\cmidrule(lr){2-4}\cmidrule(lr){5-7}\cmidrule(lr){8-10}\cmidrule(lr){11-13}
  & RNRO & LNRO & OCL & RNRO & LNRO & OCL& RNRO & LNRO & OCL& \rtwo{RNRO} & \rtwo{LNRO} & \rtwo{OCL}\\
 \midrule
NS &0.3308 & 0.3281 & \textbf{0.3573} &0.0734 &0.0705 & \textbf{0.1015}&0.1098 & 0.1068& \textbf{0.1492 }& \rtwo{0.7954} & \rtwo{0.7954} & \rtwo{\textbf{0.7796}} \\
AP &0.5984 & 0.6074 & \textbf{0.6222} & 0.0315& 0.0385& \textbf{0.0524}&0.0619 & 0.0791 & \textbf{0.0955}& \rtwo{0.5737} & \rtwo{0.5726} & \rtwo{\textbf{0.5712}}  \\
CS & 0.6225 & \textbf{0.6350} &0.6313 & 0.0545 & \textbf{0.0647} & 0.0627& 0.0726 &0.0840 & \textbf{0.0808}& \rtwo{\textbf{0.6618}} & \rtwo{0.6776} & \rtwo{ 0.6776}  \\
HR & 0.3902 & 0.3985& \textbf{0.4326}& 0.0066 &0.0131 & \textbf{0.0368} & 0.0189 &  0.0230& \textbf{0.0456}& \rtwo{0.6994} & \rtwo{0.6965} & \rtwo{\textbf{0.6946 }}\\
BC & 0.5276 &0.5476 & \textbf{0.6650}&0.0062 &0.0156 & \textbf{0.0799} & 0.0047 & 0.0083& \textbf{0.0281} & \rtwo{0.6801} & \rtwo{0.6771} & \rtwo{\textbf{0.6573}} \\
LG & 0.4493&0.4818 & \textbf{0.5426} &0.1076 & 0.1122& \textbf{0.1552} & 0.1610 & 0.1500& \textbf{0.1919}& \rtwo{0.5797} & \rtwo{\textbf{0.5733}} & \rtwo{0.5781}  \\
CR &  0.5533& 0.5583 & \textbf{0.5590}&  0.0129& 0.0141& \textbf{0.0145}& 0.0076 & 0.0082& \textbf{0.0083}& \rtwo{0.7835}  & \rtwo{0.7795} & \rtwo{\textbf{0.7777}} \\
OB &  \rtwo{0.3786}& \rtwo{0.3586}& \rtwo{\textbf{0.3935}}& \rtwo{0.1648}& \rtwo{0.1487}& \rtwo{\textbf{0.1696}}& \rtwo{0.2363} & \rtwo{0.2239}& \rtwo{\textbf{0.2633}}&\rtwo{0.3165} & \rtwo{0.3248} & \rtwo{\textbf{0.2973}}  \\
BM &  \rtwo{0.5882}& \rtwo{0.5970}& \rtwo{\textbf{0.6127}} & \rtwo{0.0223} & \rtwo{\textbf{0.0239}} & \rtwo{0.0233}&  \rtwo{0.0178}& \rtwo{ 0.0179 }& \rtwo{\textbf{0.0194}} & \rtwo{0.5817} & \rtwo{0.5805}  & \rtwo{\textbf{0.5762}}   \\
\midrule
$\overline{\textit{AR}}$ & \rtwo{2.78} & \rtwo{2.22} &  \rtwo{\textbf{ 1.11}}&  \rtwo{2.83}&  \rtwo{2.11}&  \rtwo{\textbf{1.22}}&  \rtwo{2.67}&  \rtwo{2.33}&  \rtwo{\textbf{1.11} }&  \rtwo{2.56}&  \rtwo{2.00}&  \rtwo{\textbf{1.22}}\\
\bottomrule
\end{tabular}}
\end{table*}

\subsection{Ablation Studies}\label{sct:ablation}


To demonstrate the effectiveness of the core components of OCL and to verify our hypothesis on the orders of categorical data attributes, several ablated variants of OCL are compared from both algorithm and data perspectives.

From the algorithm aspect, to evaluate the effectiveness of the proposed Order Distance (OD), we compare OCL with its variant OCL-I, which uses normalized equidistant order distance without considering the probability distribution of possible values. To evaluate the effectiveness of the Order Learning (OL) mechanism, we further remove the process of order learning from OCL-I and only update the order once in the first iteration, which forms OCL-II. To completely discard Order Information (OI), we further make OCL-II use only the traditional Hamming distance, thus forming OCL-III. Table \textcolor{blue}{\ref{tb:ab_ca}} compares the three OCL variants. It can be observed that the performance of OCL is superior to all variants. More specifically, OCL outperforms OCL-I on 17, 14, 17, and 17 out of 20 datasets in terms of CA, ARI, NMI, and CMP, respectively, indicating that OD effectively leverages the distribution of attribute values across clusters for more reasonable order distance measurement. Furthermore, OCL-I outperforms OCL-II on 17, 16, 15, and 13 out of 20 datasets in terms of CA, ARI, NMI, and CMP, respectively, verifying the effectiveness of the OL mechanism in searching for optimal orders during clustering. In addition, compared to OCL-III, which completely disregards OI, OCL-II achieves better performance on 16, 18, 17, and 19 out of 20 datasets in terms of CA, ARI, NMI, and CMP, respectively, demonstrating that even a roughly estimated initial order can effectively improve clustering performance.

From the data aspect, to validate our hypothesis that correct order information is crucial to categorical data clustering, ablation experiments are conducted on datasets containing both nominal and ordinal attributes. The OCL that learns order for both nominal and ordinal attributes is compared with its two variants: (1) LNRO (Learn Nominal Reserve Ordinal), which retains the semantic order of the original ordinal attribute and only learns orders for the nominal attributes, and (2) RNRO (Reserve Nominal Reserve Ordinal), which preserves the original semantic order for ordinal attributes without learning, and similarly imposes no order learning for nominal attributes. Table~\textcolor{blue}{\ref{tb:ab_ca_d}} compares the clustering performance of the three variants in CA, ARI, and NMI indices. It can be observed that, in terms of CA and ARI, on seven out of the nine datasets, LNRO outperforms RNRO. This verifies our hypothesis that the nominal attributes also contain potential order information that is effective for clustering. AS for the NMI and CMP indices, LNRO outperforms RNRO on six out of nine datasets, the performance gap between LNRO and RNRO is tiny, which can still verify the effectiveness of learning orders for nominal attributes in general. Moreover, we can also find that the OCL outperforms LNRO on eight out of the nine datasets in CA and ARi indices, and seven out of the nine datasets in NMI and CMP indices. This indicates that the orders learned by the proposed mechanism for ordinal attributes are more conducive to clustering than the original semantic orders.

\subsection{Convergence and Efficiency Evaluation}
\label{sct:converge_time}
\begin{figure}[!t]	
\centerline{\includegraphics[width=3.2in]{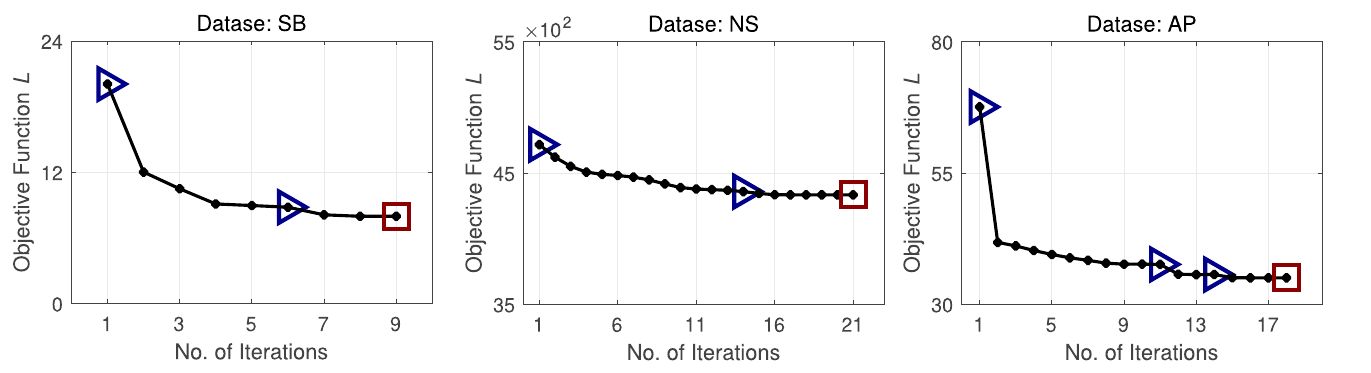}}
\caption{Convergence curves of OCL on three datasets.}
\label{fig:conv_curve_all}	
\end{figure}

The convergence of OCL is evaluated by executing it on the SB, NS, and AP datasets, and present the convergence curves in Figure~\textcolor{blue}{\ref{fig:conv_curve_all}}. The horizontal and vertical axes represent the number of learning iterations and the value of the objective function $L$, respectively. The blue triangles mark the iterations of order updates, while the red square indicates the iteration that OCL converges. It can be observed that, after each order update, the value of $L$ further decreases. This indicates that the proposed order learning mechanism is consistent with the optimization of the clustering objective function. Moreover, for all the tested datasets, OCL completes the learning process within 40 iterations with at most three updates of the orders. This reflects that OCL converges efficiently with guarantee, which conforms to the analysis in Section~\textcolor{blue}{\ref{sct:theoretical_analysis}}.

\begin{figure}[!t]	
\centerline{\includegraphics[width=3.2in]{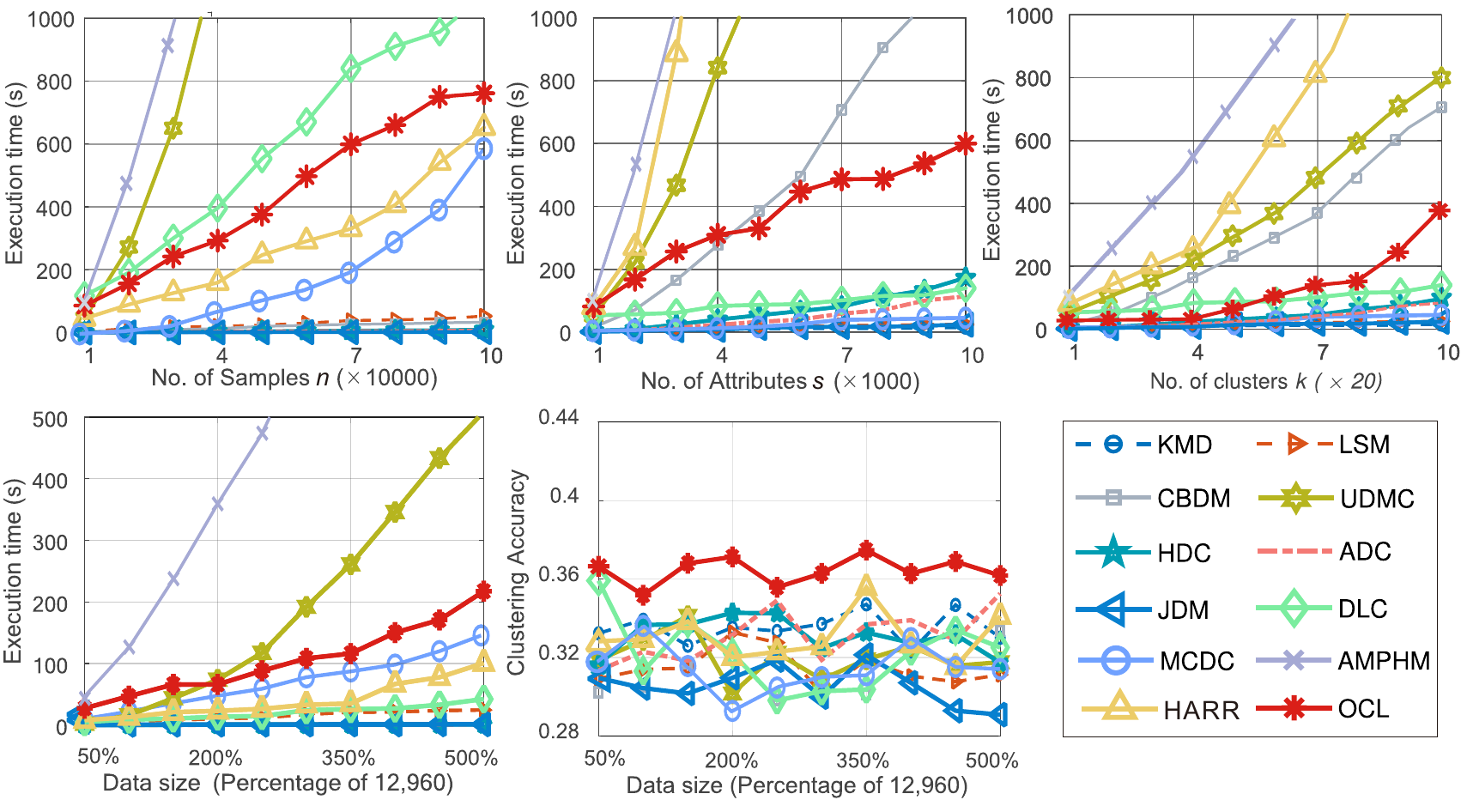}}
\caption{\rone{Efficiency evaluation on synthetic datasets (varying $n$, $s$, $k$) and real NS dataset (varying $n$, accuracy reported).}}	
\label{fig:time_complex}	
\end{figure}


\rone{To evaluate the efficiency of OCL, large synthetic datasets are generated with varying numbers of samples ($n$), attributes ($s$), and clusters ($k$) with the following configurations: (1) fix $s = 20$ and $k = 5$, and vary $n$ from 10k to 100k in steps of 10k, (2) fix $n = 2k$ and $k = 5$, and vary $s$ from 1k to 10k in steps of 1k, and \mrone{(3) fix $n = 2k$ and $s = 20$, and vary $k$ from 20 to 200 in steps of 20.} All sample values are randomly generated from five possible values per attribute. Overall, Figure~\textcolor{blue}{\ref{fig:time_complex}} 
shows that OCL achieves higher efficiency than state-of-the-art methods while maintaining comparable scalability to traditional baseline methods. More specifically, on the synthetic datasets, OCL maintains a nearly linear growth rate in execution time, which is consistent with the time complexity analysis in Theorem~\textcolor{blue}{\ref{the:time_analysis}}, and its efficiency outperforms:
(1) UDMC, AMPHM, and DLC on large-scale datasets;
(2) HARR, UDMC, AMPHM, and CBDM on high-dimensional datasets; \mrone{(3) the same competitors on datasets with numerous clusters.} In addition, the evaluation is also conducted on the real benchmark NS dataset with varying sampling rates (from 50\% to 500\% in steps of 50\%) to show that OCL achieves scalability without compromising clustering accuracy. In summary, OCL is effective and efficient compared to state-of-the-art methods, without incurring much additional computational cost in comparison with the traditional simple approaches.}

\begin{figure}[!t]	
\subfigure{
\begin{minipage}{0.27\linewidth}	
  \centerline{\includegraphics[width=1in]{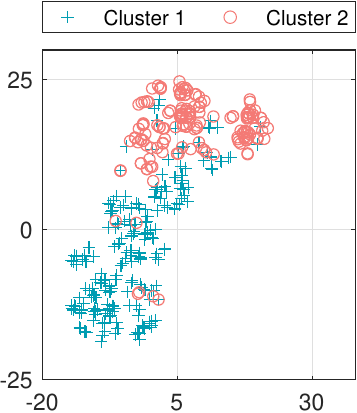}}	
\end{minipage}	
}
\hfill	
\subfigure{
\begin{minipage}{0.27\linewidth}	
  \centerline{\includegraphics[width=1in]{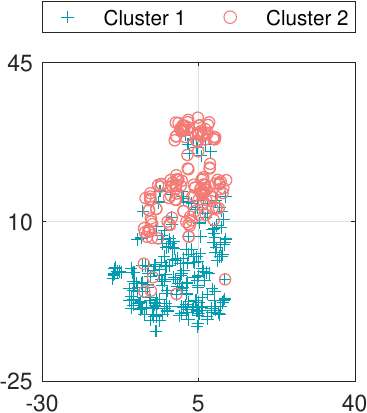}}	
\end{minipage}	
}
\hfill	
\subfigure{
\begin{minipage}{0.27\linewidth}	
  \centerline{\includegraphics[width=1in]{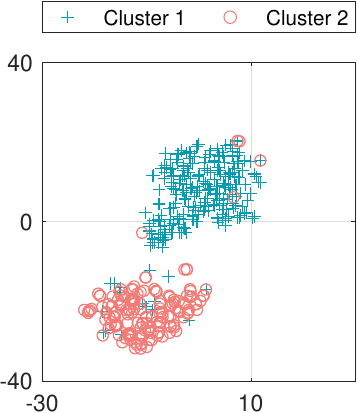}}	
\end{minipage}	
}
\caption{t-SNE visual comparison of CBDM (left), ADC (middle), and OCL (right) on the VT dataset.}	
\label{fig:Visualization_tsne_Vt}	
\end{figure}

\begin{table}[!t]
\caption{Clustering performance of mixed data clustering approaches and their OCL-enhanced versions i.e., adopting the distance metric learned by OCL on AP, AC, and DS datasets. The green arrow indicates that an improvement has been achieved by the corresponding OCL-enhanced version.}
\centering
\resizebox{0.8\columnwidth}{!}{
\begin{tabular}{c|c|lll}
\toprule
\multirow{2}{*}{\textbf{Datasets}}  & \multirow{2}{*}{\textbf{Methods}} & \multicolumn{3}{c}{\textbf{Indices}} \\ \cmidrule{3-5} 
 &  & \textbf{CA} & \textbf{ARI} & \textbf{NMI} \\ \midrule
\multirow{6}{*}{AP} & KMS+OCL &  0.5550\textcolor{darkgreen}{$\uparrow$} &  0.0104\textcolor{darkgreen}{$\uparrow$} &  0.0106\textcolor{darkgreen}{$\uparrow$} \\
 & KPT & 0.5323 & 0.0015 & 0.0051 \\ \cmidrule{2-5} 
 & HARR+OCL & 0.6042\textcolor{darkgreen}{$\uparrow$} &  0.0373\textcolor{darkgreen}{$\uparrow$} &  0.0593\textcolor{darkgreen}{$\uparrow$} \\ 
 & HARR & 0.5688 & 0.0164 & 0.0371 \\ \cmidrule{2-5} 
 & ADC+OCL &  0.5677\textcolor{darkgreen}{$\uparrow$} &  0.0144\textcolor{darkgreen}{$\uparrow$} &  0.0126\textcolor{darkgreen}{$\uparrow$} \\ 
 & ADC & 0.5487 & 0.0049 & 0.0070 \\ \midrule
\multirow{6}{*}{AC} & KMS+OCL &  0.7975\textcolor{darkgreen}{$\uparrow$} &  0.3537\textcolor{darkgreen}{$\uparrow$} &  0.2738\textcolor{darkgreen}{$\uparrow$} \\ 
 & KPT & 0.7425 & 0.2794 & 0.2154 \\ \cmidrule{2-5} 
 & HARR+OCL &  0.8232\textcolor{darkgreen}{$\uparrow$}  &  0.4169\textcolor{darkgreen}{$\uparrow$}  &  0.3225\textcolor{darkgreen}{$\uparrow$}  \\ 
 & HARR & 0.8213 & 0.4121 & 0.3184 \\ \cmidrule{2-5} 
 & ADC+OCL &  0.6480\textcolor{darkgreen}{$\uparrow$}  &  0.0839\textcolor{darkgreen}{$\uparrow$}  &  0.0644\textcolor{darkgreen}{$\uparrow$}  \\ 
 & ADC & 0.6445 & 0.0797 & 0.0607 \\ \midrule
\multirow{6}{*}{DS} & KMS+OCL & 0.6842\textcolor{gray}{-} & 0.1684\textcolor{gray}{-} & 0.1888\textcolor{gray}{-} \\ 
 & KPT & 0.6842 & 0.1684 & 0.1888 \\ \cmidrule{2-5} 
 & HARR+OCL & 0.8117\textcolor{gray}{-}	&0.4364\textcolor{gray}{-}	&0.4277\textcolor{gray}{-} \\ 
 & HARR & 0.8117	&0.4364	&0.4277\\ \cmidrule{2-5} 
 & ADC+OCL  & 0.8667\textcolor{gray}{-}	&0.5336\textcolor{gray}{-}	&0.4244\textcolor{gray}{-}
 \\ 
 & ADC & 0.8667	&0.5336	&0.4244 \\ 
 \bottomrule
\end{tabular} }
\label{table:mixed_p}
\end{table}

\subsection{Visualization of Clustering Effect}
\label{sct:v_c_study}
To demonstrate the cluster discrimination ability of the OCL method, We use OCL, CBDM, and ADC to learn a distance metric for the VT dataset, then compute a new distance matrix for the data samples, which are projected into 2-D space for visualization via t-SNE~\textcolor{blue}{\cite{r1visual}}, as shown in Figure~\textcolor{blue}{\ref{fig:Visualization_tsne_Vt}}. We use different markers (distinguished in colors and shapes) to indicate ground-truth clusters, with point positions representing the samples projected into the 2-D space by t-SNE. This effectively visualizes how well each method captures the distinguishable cluster distributions for intuitive comparison. It can be seen from Figure \textcolor{blue}{\ref{fig:Visualization_tsne_Vt}} that OCL has significantly better cluster discrimination ability, as it performs clustering task-oriented distance learning to better suit the exploration of the $k^*$ clusters. 

\subsection{Clustering Performance on Mixed Datasets}
~\label{AP:CPMD}
To demonstrate the potential of OCL in boosting the clustering of mixed data comprising both numerical and categorical attributes, we compare the original $k$-ProTotypes(KPT)~\textcolor{blue}{\cite{kpt}}, HARR~\textcolor{blue}{\cite{HARR}}, and ADC~\textcolor{blue}{\cite{ADC}} methods proposed for mixed data clustering, with their corresponding variants enhanced by the orders learned by OCL. More specifically, for KPT, we first encode the categorical attribute values using the orders learned by the proposed OCL method and then treat the dataset as a pure numerical dataset for clustering using the $k$-means (KMS) algorithm~\textcolor{blue}{\cite{kms}}. Actually, KPT enhanced by OCL is equivalent to KMS with the categorical attributes encoded into numerical ones by OCL, and thus we call the enhanced version KMS+OCL. For HARR and ADC, since they are originally competent in processing ordinal attributes, the orders learned by OCL are assigned to the categorical attributes, and we let HARR and ADC treat all the categorical attributes as ordinal attributes. Accordingly, their enhanced versions are named HARR+OCL and ADC+OCL, respectively. Table~\textcolor{blue}{\ref{table:mixed_p}} demonstrates the performance of the above-mentioned approaches and their OCL-enhanced versions.

It can be observed that OCL considerably enhances the clustering performance of the three approaches in terms of AP and AC datasets. This significantly indicates that the learned orders can effectively improve the clustering performance on mixed data. The improvement on the AP dataset is more obvious because the AP contains both nominal and ordinal attributes as shown in Table~\textcolor{blue}{\ref{tb:detaile_infor}}. This situation allows OCL to leverage its superiority in learning orders for nominal attributes and potential corrections on the semantic orders of ordinal attributes. A noteworthy special case can be seen on the DS dataset. Since DS has only two possible values on each of its attributes, OCL cannot further optimize the orders. This is why the clustering performance on the DS dataset remains unchanged after using the orders learned by OCL. 
This observation suggests that when the original semantic order of an ordinal attribute is already optimal, OCL exerts no measurable effect on clustering performance. In short, the proposed OCL is promising for more complex mixed-data clustering scenarios.

\begin{figure}[!t]	
\centerline{\includegraphics[width=3.2in]{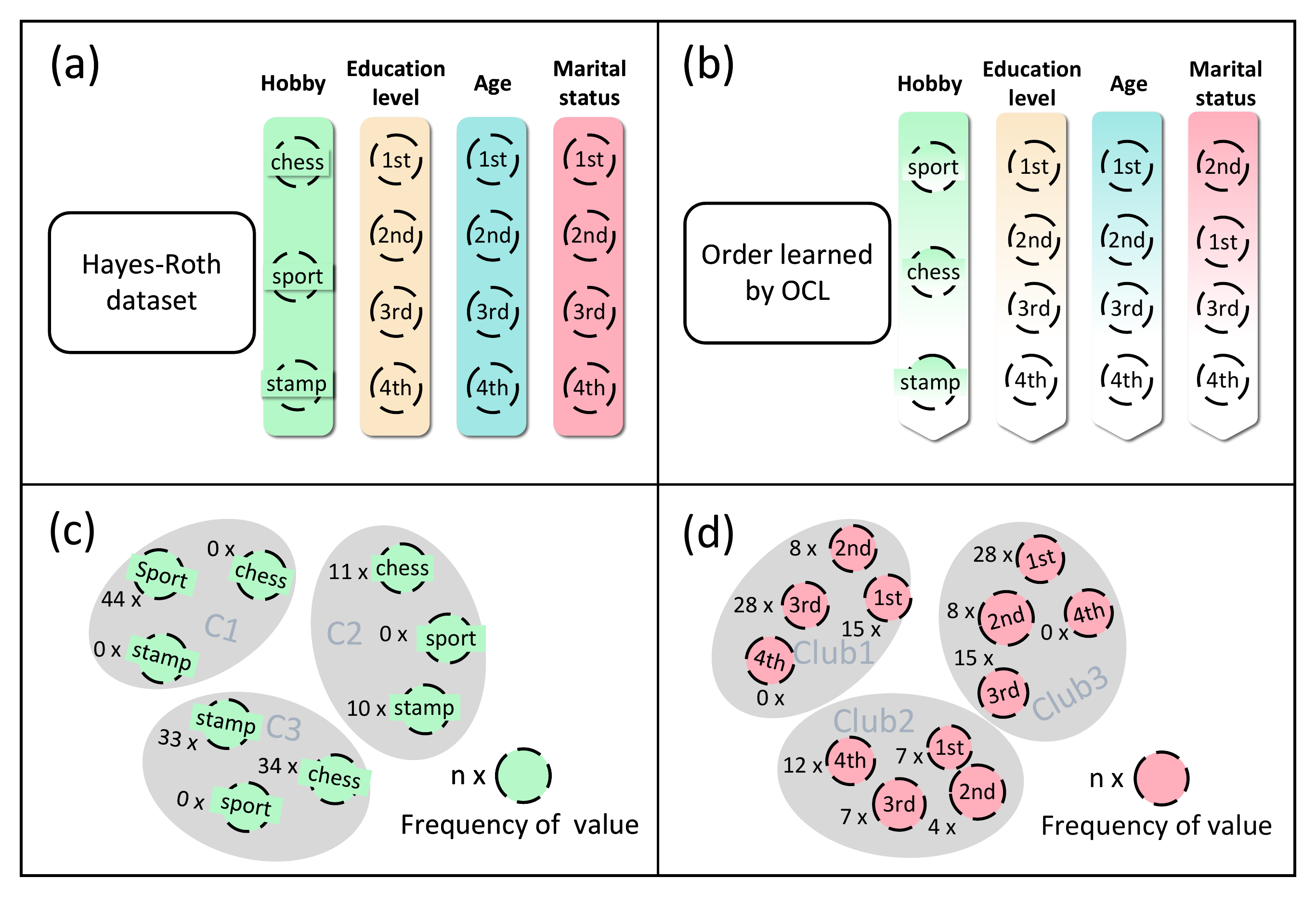}}
\caption{A case study on HR dataset. ``Hobby'' is a nominal attribute with possible values $\{$chess, sport, stamp$\}$, and the remaining attributes are ordinal attributes with the semantic order of their possible values indicated by $\{$1st, 2nd, 3rd, 4th$\}$. Sub-figure (a) demonstrates the original data. Sub-figure (b) showcases the orders learned by the OCL method. Sub-figure (c) illustrates the possible values of the ``Hobby'' attribute in the clusters obtained by OCL. Sub-figure (d) presents the possible values of the ``Marital status'' attribute in the true clusters indicated by the data labels.}	
\label{fig:case_study}	
\end{figure}

\subsection{Case Study on Hayes-Roth (HR) Dataset}
\label{AP: CS}

It is worth noting that the proposed order learning method not only improves the clustering performance but also mines the hidden order information when facing datasets with unclear semantic information. This case study investigates whether the value order learned by the proposed method is interpretable and consistent with human subjective understanding. The HR dataset with one nominal and three ordinal attributes records personal information, such as the nominal ``Hobby'' and the ordinal ``Education level'', ``Age'', and ``Marital status''. The true labels divide the samples into three groups of club members for psychological evaluation. The original dataset information is visualized in Figure~\textcolor{blue}{\ref{fig:case_study}} (a).

It can be seen from Figure~\textcolor{blue}{\ref{fig:case_study}} (b) that OCL can correctly learn the semantic order of the values from attributes ``Education level'' and ``Age'', as the learned orders are consistent with the original semantic orders. Moreover, for the nominal attribute ``Hobby'', although there is no original obvious semantic order, the learned order is very meaningful in data analysis and knowledge acquisition. First, the learned order distance between ``chess'' and ``stamp collection'' hobbies is closer, while the distance between ``stamp collection'' and ``sport'' is larger. This is consistent with cognitive intuition that ``chess'' is a quieter sport, so it ranks between the quieter hobby ``stamp collection'' and its sibling ``sport''. This indicates that OCL can learn latent semantic orders for nominal attributes, which is an important characteristic for categorical data understanding.

As shown in the clustering results obtained by OCL in Figure~\textcolor{blue}{\ref{fig:case_study}} (c), it can be seen that there are more cases of people who enjoy playing chess and collecting stamps appearing in the same cluster, which is consistent with human intuition and the learned order of ``Hobby'' attribute. This further confirms the prospects of our OCL method in discovering the implicit order information of unlabeled data with no obvious semantic order. 

For the ``Marital status'' attribute, the order learned by OCL is inconsistent with the original semantic order as demonstrated in Figure~\textcolor{blue}{\ref{fig:case_study}} (b). The order learned by OCL indicates that the possible values semantically ranked first and third in ``Marital status'' should be more similar to better serve clustering. From the true clusters given by the dataset shown in Figure~\textcolor{blue}{\ref{fig:case_study}} (d), it can be found that more of the first and third values of the ``Marital status'' attribute simultaneously appear in the same clusters, which confirms the reasonableness of the learned order. This characteristic allows the users to better understand the cluster distribution of data, or even explore deeper meanings and relationships of values.

\section{Concluding Remarks}
This paper introduces and validates a new finding that knowing the correct order relation of attribute values is critical to cluster analysis of categorical data. Accordingly, the OCL paradigm is proposed, allowing joint learning of data partitions and order relation-guided distances.
Its learning mechanism can infer a more appropriate order among attribute values based on the current clusters, thus yielding more compact clusters based on the updated order. The data partition and order learning are iteratively executed to approach the optimal clustering solution. As a result, more accurate clustering results and insightful orders can be obtained.
OCL converges quickly with a guarantee, and extensive experiments on 20 real benchmark datasets fully demonstrate its superiority in comparison with the state-of-the-art clustering methods. A case study has also been provided to show the rationality of the learned order in understanding categorical data clusters.
The proposed OCL is also not exempt from limitations. For instance, the natural connection between the treatment of categorical and numerical attributes remains to be explored. Moreover, it would be promising to extend OCL into more complex scenarios, e.g., processing non-stationary mixed dataset with unknown number of imbalanced clusters.

\section*{ACKNOWLEDGEMENTS}
This work was supported in part by the National Natural Science Foundation of China (NSFC) under grants: 62476063, 62376233, 61806131, and 62306181, the NSFC/Research Grants Council (RGC) Joint Research Scheme under the grant N\_HKBU214/21, the Natural Science Foundation of Guangdong Province under grant: 2025A1515011293, the Natural Science Foundation of Fujian Province under grant: 2024J09001, the National Key Laboratory of Radar Signal Processing under grant: JKW202403, the General Research Fund of RGC under grants: 12201321, 12202622, and 12201323, the RGC Senior Research Fellow Scheme under grant: SRFS2324-2S02, the Shenzhen Science and Technology Program under grant: RCBS20231211090659101, the Guangdong Provincial Key Laboratory under grant: 2023B1212060076, and the Xiaomi Young Talents Program.
\bibliographystyle{ACM-Reference-Format}
\bibliography{SigMOD26_olc}

\end{document}